%% file: main.tex
\documentclass[journal,twoside,web]{ieeecolor}
\usepackage{amsmath,amsfonts} 
\usepackage{array}
\usepackage[caption=false,font=normalsize,labelfont=sf,textfont=sf]{subfig}
\usepackage{textcomp}
\usepackage{stfloats}
\usepackage{url}
\usepackage{verbatim}
\usepackage{graphicx}
\usepackage{bm}
\usepackage{booktabs}
\usepackage{generic}
\usepackage[noadjust]{cite}

\usepackage{amsmath,amssymb,amsfonts,amsthm}
\usepackage{algorithm}
\usepackage{algorithmic}
\usepackage{wasysym}
\usepackage{hyperref}
\usepackage{mathrsfs}
\usepackage{graphicx,float}
\usepackage{capt-of}
\usepackage{textcomp,soul,comment}
\usepackage{pgfplots}
\usepackage{mathtools}

\pgfplotsset{compat=1.10}

\usepackage{cleveref}

\usepackage[detect-all]{siunitx}

\newtheorem{theorem}{Theorem}
\newtheorem{proposition}{Proposition}
\newtheorem{assumption}{Assumption}
\newtheorem{lemma}{Lemma}

\newtheorem{remark}{Remark}

\crefname{assumption}{Assumption}{Assumptions}
\crefname{prop}{Proposition}{Propositions}

\newcommand{\B}[1]{\mathbb{#1}}
\newcommand{\C}[1]{\mathcal{#1}}
\newcommand{\vertex}[1]{\mathrm{Vert}\left(#1\right)}
\newcommand{\dist}[1]{\mathrm{dist}\left(#1\right)}

\def\revised{\textcolor{black}}

\title{An MPC framework for efficient navigation of mobile robots in cluttered environments}
\author{Johannes K\"ohler$^\star$, Daniel Zhang$^\star$, Raffaele Soloperto, Andrea Carron, Melanie Zeilinger%
\thanks{$\star$The first two authors contributed equally to this work}
\thanks{Johannes Köhler, Andrea Carron, and Melanie Zeilinger are with the Institute for Dynamic Systems and Control, ETH Zurich, Zurich CH-8092 (e-mail: \{jkoehle,carrona,mzeilinger\}@ethz.ch). 
Johannes Köhler is now with the Department of Mechanical Engineering, Imperial College London (e-mail: j.kohler@imperial.ac.uk). 
Daniel Zhang is a student at ETH Zurich, Zurich CH-8092, (email: danzhang@student.ethz.ch). 
Raffaele Soloperto is with the Automatic Control Laboratory, ETH Zurich, Zurich (email: soloperr@ethz.ch).}%
\thanks{Johannes K\"ohler was supported by the Swiss National Science Foundation under NCCR Automation (grant agreement 51NF40 180545).}
}

\begin{document}
\maketitle

\begin{abstract}
We present a model predictive control (MPC) framework for efficient navigation of mobile robots in cluttered environments. The proposed approach integrates a finite-segment shortest path planner into the finite-horizon trajectory optimization of the MPC. This formulation ensures convergence to dynamically selected targets and guarantees collision avoidance, even under general nonlinear dynamics and cluttered environments. 
The approach is validated through hardware experiments on a small ground robot, where a human operator dynamically assigns target locations \revised{ that a robot should reach while avoiding obstacles.  The robot reached new targets within 2-3 seconds and responded to new commands within \SIrange{50}{100}{\milli\second}, immediately adjusting its motion even while still moving at high speeds toward a previous target.} 
\end{abstract}

\section{Introduction}
\IEEEPARstart{A}{utonomous} mobile robots are increasingly playing an essential role in a wide range of applications~\cite{siegwart2011introduction}, such as warehouse logistics~\cite{wurman2008coordinating}, industrial inspection~\cite{gehring2021anymal}, and factory automation~\cite{tilley2017automation}. 
These tasks rely on efficient navigation through complex environments, requiring reliable motion planning to avoid collisions. 
There is a growing demand for mobile robots to be more responsive, quickly accomplish such challenging navigation tasks and adapt to dynamically changing obstacles and user commands~\cite{tordesillas2021faster,hoy2015algorithms}. 
This is, however, challenging to achieve with established control methods, which are based on rigid hierarchical structures~\cite{LaValle2006combinatorial,zhou2022review}.

\begin{figure}[t]
\centering
\includegraphics[width=0.45\textwidth]{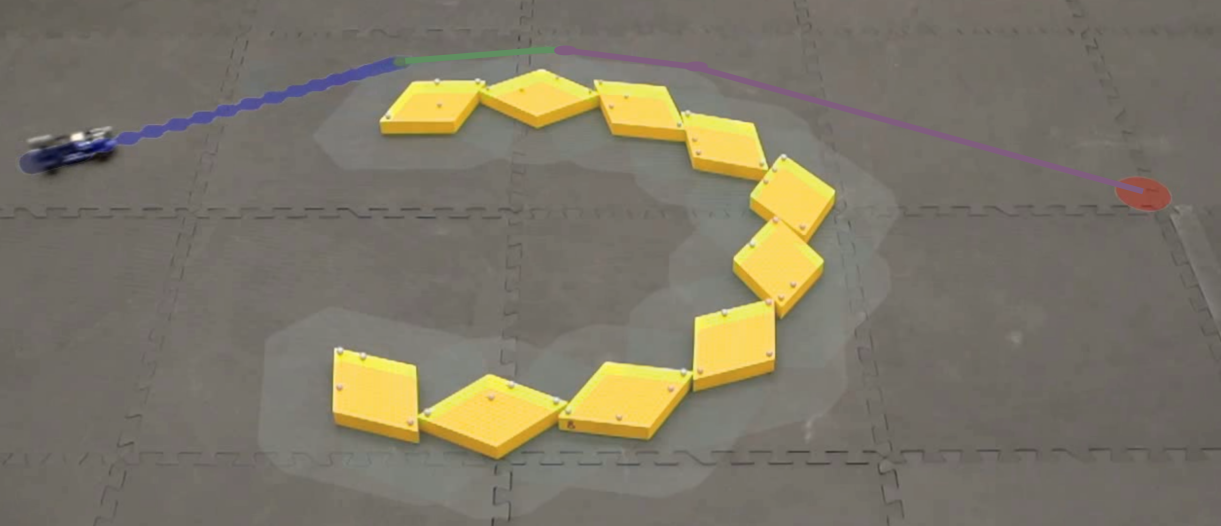}
\caption{Visualization of the proposed control approach to navigate through cluttered environments.  
The MPC formulation jointly optimizes a short dynamic trajectory (blue) followed by a discrete path (green, purple) to reach the user-specified target (red circle), while avoiding obstacles (yellow).} 
\label{fig:page1}
\end{figure}

\subsubsection*{State-of-practice}
Motion planning of mobile robots is classically decomposed into multiple modules~\cite{zhou2022review}. Global path-planners, like $A^\star$ or RRT, generate a path to the desired target~\cite{LaValle2006combinatorial}. Then, optimization-based techniques, like model predictive control (MPC)~\cite{rawlings2017model}, generate feasible trajectories, taking into account the dynamics and constraints of the robot, see, e.g.,~\cite{brito2019model,wullt2025robust,zhang2025integrating,ghorab2025multi} for recent works. 
This approach relies on a time scale separation between planning, trajectory optimization, and control, such that their interaction can be neglected in the design~\cite{matni2024quantitative}. 
\revised{As a result, collision-free motion can be reliably generated, however, obstacles and desired behaviour are only addressed on the highest layer and are difficult to update on-the-fly while a robot is executing a mission.}\color{black}{ } 
However, in dynamic and uncertain environments, especially those involving human–robot interaction, there is a growing need for robots to adapt and re-plan in real time based on new commands and information. 
\revised{In this work, we address this gap by developing approaches that retain the reliability and safety of classical methods while enabling rapid response to new user commands.}

\subsection*{Related work}
\subsubsection*{MPC for tracking}
\begin{figure}
    \centering
Standard approach \quad \quad\quad\quad \quad \quad Proposed approach\vspace{2mm}\\    \includegraphics[width=\linewidth]{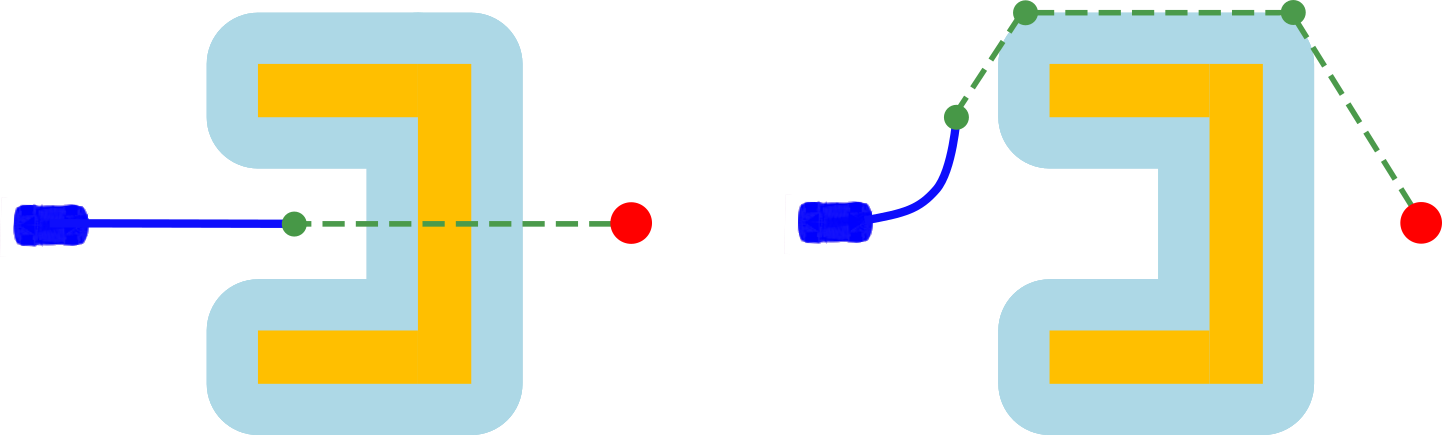}
\caption{Illustration of a standard MPC for tracking~\cite{limon2018nonlinear,krupa2024model} (left) and the proposed formulation (right) in an environment obstructed by a challenging obstacle. 
Obstacles (orange) are surrounded by a buffer to account for robot geometry (light blue region). The mobile robot (blue) tries to reach the target (red circle). The MPC optimizes a trajectory (blue) to reach an artificial reference (green circle) and minimize an offset cost to the target. 
Left: The  distance is directly minimized and the robot gets stuck \revised{in front of the obstacle}.
Right: The offset cost optimizes a 3-segment path (green, dashed) to the target and successfully navigates around the obstacle towards the target.}
\label{fig:illustrate_offset_stuck}
\end{figure}

Reference governors~\cite{garone2017reference} are a classical strategy for managing constraints and dynamic targets by computing a separate (artificial) reference that can be safely passed to a controller. 
\revised{This provides a lightweight solution for ensuring collision-free navigation~\cite{hermand2018constrained}. To further enhance performance, }this concept can be tightly integrated into an MPC formulation, yielding \emph{MPC for tracking}~\cite{limon2008mpc}. 
These MPC formulations jointly optimize a prediction of the dynamical system over a finite horizon and the artificial reference, where an offset cost penalizes the difference between the artificial reference and the target. 
This ensures that feasibility and safety is guaranteed independent of changes in the target. 
A detailed discussion on the benefits of MPC for tracking approaches can be found in the overview paper~\cite[Chap.~4]{kohler2024analysis} and the tutorial~\cite{krupa2024model}. 
Asymptotic stability guarantees for these approaches \revised{rely on a
convexity assumption (cf. \cite[Asm.~2]{limon2018nonlinear})} to ensure that the artificial reference will be incremented towards the target. 
In cluttered environments with many obstacles, this convexity condition does not hold and these approaches can instead get stuck, failing to reach the target. 
\revised{This issue is also visualized in Figure~\ref{fig:illustrate_offset_stuck} and we also demonstrate the limitations of this approach in the numerical comparison later.} The challenges associated with non-convex environments  have been investigated in detail in~\cite{cotorruelo2020tracking,dos2024stability,soloperto2023safe,soloperto2022nonlinear}.
In~\cite{cotorruelo2020tracking}, it is established that non-convex sets in normal form can be addressed by designing a tailored non-convex offset cost. However, applicability of this methodology to cluttered environments is unclear. 
In~\cite{dos2024stability}, obstacle avoidance is addressed through a soft penalty, however, convergence to the target cannot be ensured. 
In~\cite{soloperto2023safe}, it is suggested that a sufficiently large offset cost can help escape local minima. However, this requires a sufficiently large prediction horizon to circumvent large obstacles and assumes that the solver obtains the global optimum, which significantly increases computational demand. 
A more systematic solution has been proposed in~\cite[Sec.~4]{soloperto2022nonlinear}: If the offset cost equals the length of the shortest feasible curve connecting the artificial reference to the target, then stability guarantees hold in general non-convex environments. 
However, this formulation is not directly tractable and, in fact, has never been successfully implemented, even in simulation. 
Our approach is inspired by the analysis in~\cite{soloperto2022nonlinear}, but differs by integrating insights from global path planners~\cite{LaValle2006combinatorial} to enable applicability to cluttered environments and under fast sampling rates.

\subsubsection*{Hierarchical tracker-planner formulations}
The computational demand of jointly optimizing a trajectory and a plan can be reduced by optimizing both separately in a tracker and planner. 
By suitably co-designing the planner and tracker, these formulations inherit the guarantees from the MPC for tracking formulation~\cite{koehler2020nonlinearAutomatica,convens2024terminal,benders2025embedded,eberhard2025time}, 
while allowing a flexible trade-off between computational demand and responsiveness.
Since the planner in~~\cite{koehler2020nonlinearAutomatica,benders2025embedded,convens2024terminal,eberhard2025time} corresponds to a simple offset cost, these approaches can also get stuck in cluttered environments without reaching the target. 
In addition, the co-design in~\cite{koehler2020nonlinearAutomatica,convens2024terminal}
crucially relies on the offline design of a large invariant/contractive terminal set, which is challenging for complex robot dynamics~\cite{koehler2020nonlinearTAC}, especially for non-holonomic wheeled robots, see also~\cite[Sec.~4.4]{kohler2024analysis} for a detailed discussion. 
In contrast, the proposed formulation can leverage simple terminal equality constraints and flexibly trade off the degrees of freedom in the planner. 

\subsubsection*{Experimental application} 
Although MPC for tracking has matured into a solid theoretical framework, applications to robotics platforms with collision avoidance are still largely limited to simulation experiments~\cite{santos2023nonlinear,convens2024terminal}, with the possible exception being manipulator experiments in~\cite{Nubert2020Robot}. 
The hierarchical approach~\cite{benders2025embedded} has been implemented in hardware experiments with a mobile drone and obstacles; however, updates in the target only change the control command after a delay of over one second, which limits responsiveness. 
In contrast, we present hardware experiments with fast-moving wheeled robots that respond to new targets within \revised{\SIrange{50}{100}{\milli\second}}.

\subsection*{\revised{Contribution}}
We propose an MPC framework that integrates tracking and planning, ensuring reliable and efficient navigation through \revised{environments cluttered with (static) obstacles}, see Figure~\ref{fig:page1}.
Specifically, we address the problem where a desired target location is specified at runtime and the robot must quickly navigate to this location while reliably avoiding obstacles.\\ 
\subsubsection*{Method}The proposed MPC formulation optimizes a finite-horizon trajectory that accounts for the shape of the robot geometry and obstacles, the nonlinear robot dynamics, and the state and input constraints, and ensures safe operation of the robot. 
In addition to the dynamic trajectory, our formulation jointly optimizes a finite-segment path from the end of the trajectory to the desired target. This path does not need to account for the dynamics, but only determines if we make progress towards the target.   
This path is efficiently initialized using the shortest path road map and Dijkstra's algorithm. 
By combining the trajectory optimization and the global planner, our method yields dynamically feasible trajectories, accounts for global long-term planning, and can react swiftly to changes in the environment.\\ 
\subsubsection*{Theoretical guarantees}The proposed method guarantees recursive feasibility, constraint satisfaction, collision avoidance, and convergence to the target. 
Due to the use of trajectory optimization and a global planner, these guarantees hold for general nonlinear robot dynamics and environments cluttered with many  \revised{(static) }obstacles. 
\revised{Notably, the receding-horizon implementation using both trajectory optimization and path optimization ensures collision-avoidance for the dynamics, even though the computed path is in general not dynamically feasible.} 
\subsubsection*{Implementation \& Experiments}We provide an open-source implementation of this method:
\begin{center}
{\small \url{https://github.com/IntelligentControlSystems/ClutteredEnvironment}}
\end{center}
We demonstrate the results in hardware experiments with an RC-car using two different environments cluttered with  \revised{(static) }obstacles. 
A human operator dynamically changed the target\revised{, even while the robot was still in high speed motion. 
The robot changed its motion within $\SIrange{50}{100}{\milli\second}$, } navigated through obstacles and reached new targets within $2$–$3$ seconds, as shown in the video: 
\begin{center}
{\small \url{https://youtu.be/Hn_hpAmGgq0}}
\end{center}
We also conducted simulation experiments in randomly generated environments, where we compared our approach to existing MPC formulations, which do not incorporate a global planner.  
While the proposed method is successful in $100\%$ of the trials, existing formulations get stuck in most (densely) cluttered environments. 
\revised{The addition of this global planner only increases the worst-case computational demand by about $17$--$35\%$ across different experiments}.

\subsubsection*{Outline}
We first introduce the problem setup in Section~\ref{sec:setup}. 
Then, we present the proposed MPC formulation (Sec.~\ref{sec:method}) and the theoretical analysis (Sec.~\ref{sec:theory}). 
Section~\ref{sec:efficient} provides a more computationally efficient algorithm and establishes that this approach inherits the safety and convergence properties. 
Experiments are detailed in Section~\ref{sec:exp} and the paper ends with a conclusion in Section~\ref{sec:conclusion}. 
All theoretical proofs can be found in the appendix.

\subsubsection*{Notation}
By $\B{N}_{a:b}$ we denote the set of integers in the interval [a,b].
For two vectors $a\in\B{R}^{n_1},b\in\B{R}^{n_2}$, we abbreviate the stacked vector as $(a,b):=[a^\top,b^\top]^\top\in\B{R}^{n_1+n_2}$.
We denote the vector of ones by $[1,\dots,1]^\top=\mathbf{1}_n\in\B{R}^n$. 
Given two vectors $y_1,y_2\in\B{R}^n$, $\overline{y_1 y_2}$ denotes the line segment connecting both, i.e., their convex hull. 
Given a set $\B{C}\subseteq \B{R}^n$, $\operatorname{int}(\B{C})\subseteq\B{R}^n$ denotes its interior. 
Given a convex polytope $\B{C}\subseteq\B{R}^n$, we denote its vertices by $\vertex{\B{C}}\in\B{R}^{n\times n_{\mathrm{vert}}}$. 
For a vector  $x\in\B{R}^n$, we denote the Euclidean norm by $\|x\|=\sqrt{x^\top x}$. 
For any $\epsilon>0$, we denote the ball $\B{B}_\epsilon=\{x\in\B{R}^n|~\|x\|\leq \epsilon\}$. Given a vector $x\in\B{R}^n$ and a set $\B{A}\subseteq\B{R}^n$, we denote the point-to-set-distance by $\|x\|_{\B{A}}=\inf_{a\in\B{A}}\|x-a\|$. 
For two sets, $\B{A},\B{C}\subseteq\B{R}^n$, we similarly define the distance as $\dist{\B{A},\B{C}}=\inf_{a\in\B{A},c\in\B{C}}\|a-c\|$, 
with $\dist{\B{A},\B{C}}=0$ if $\B{A}\cap\B{C}\neq\emptyset$. 
For two sets $\B{A},\B{C}\subseteq\B{R}^n$, $\B{A}\oplus\B{C}=\{a+c \mid a\in\B{A},c\in\B{C}\}$ denotes the Minkowski sum. 
A function $\alpha$ is of class $\mathcal{K}$, i.e., $\alpha\in\mathcal{K}$, if $\alpha:\B{R}_{\geq 0}\rightarrow \B{R}_{\geq 0}$, $\alpha$ is continuous, strictly increasing, and $\alpha(0)=0$. 
A function $\alpha\in\mathcal{K}$ is of class $\mathcal{K}_\infty$, i.e., $\alpha\in\mathcal{K}_\infty$, if it is additionally radially unbounded\revised{, i.e., $\lim_{r\rightarrow\infty}\alpha(r)=\infty$}.  
A function $\beta:\B{R}_{\geq 0}\times \B{N}\rightarrow\B{R}_{\geq 0}$ is of class  $\C{KL}$, i.e., $\beta\in\C{KL}$, if $\beta$ is continuous,  for any $t\in\B{N}$: $\beta(\cdot,t)\in\mathcal{K}$, and $\forall  r\in\B{R}_{\geq 0}$: $\lim_{t\rightarrow\infty} \beta(r,t)=0$ .

\section{Problem setup}
\label{sec:setup}
We consider a mobile robot with nonlinear dynamics
\begin{align}
    x(t+1) &= f(x(t), u(t)) \label{eq:dynamics}
\end{align}
where $x(t)\in\B{R}^n$ is the state, $u(t)\in\B{R}^m$ is the control input, and $t\in\B{N}$ is the discrete time. 
The state is assumed to be measured and the dynamics $f$ are known. 
The system is subject to state and input constraints 
\begin{align}
\label{eq:constraints}    
(x(t),u(t))\in\B{Z},\quad \forall t\in\B{N},
\end{align}
reflecting, e.g., actuator limitations and bounds on the maximal velocity. 
Let $\B{Y}:=\B{R}^p$ denote the Cartesian space in which the robot operates, i.e., $p\in\{2,3\}$ for $2$D or $3$D environments. 
The occupied space of the robot $\C{Y}(t)\subseteq \B{Y}$ and the position of the robot $y(t)\in\B{Y}$ are given by known nonlinear functions 
\begin{align}
\label{eq:output}
 y(t) = & h(x(t)),\quad 
 \C{Y}(t) = \C{H}(x(t)).
\end{align}
The robot geometry is over-approximated by the convex polytope $\C{H}(x)$ with vertices $\vertex{\C{H}(x)}\in\B{R}^{p\times n_{\mathrm{vert},\C{H}}}$ $\forall x\in\B{R}^n$. 
Typically, we have $\C{H}(x)=\{h(x)\}\oplus \C{R}(x)\C{H}_0$, where the matrix $\C{R}(x)$ accounts for rotations and $\C{H}_0$ is a fixed polytopic over-approximation of the shape of the robot. 
There are \revised{known (static) }polytopic obstacles $\B{O}_i \subseteq \B{Y}$, $i\in\B{N}_{1:n_{\mathrm{o}}}$,  which need to be avoided. 
Specifically, we need to ensure 
\begin{align}
\label{eq:collision_avoidance}    
\dist{\B{O}_i,\C{H}(x(t))}\geq \delta_{\mathrm{obst}},~\forall t\in\B{N},~i\in\B{N}_{1:n_{\mathrm{o}}},
\end{align}
with a distance $\delta_{\mathrm{obst}}>0$. 
Thus, the combined input, state, and collision avoidance constraints are given by
\begin{align} \label{eq:obstacle_constraint_set}
    \B{Z}_{\B{O}} :=& \{(x, u) \in \B{Z} \mid  \dist{\C{H}(x),\B{O}_i}\geq \delta_{\mathrm{obst}},~i\in\B{N}_{1:n_{\mathrm{o}}}\}.
\end{align} 
During system operation, we are provided with a target set $\B{Y}_t^{\mathrm{d}}\subseteq \B{Y}$ \revised{ and the goal is to ensure that the distance between the robot position $y(t)$ and the target $\B{Y}_t^{\mathrm{d}}$ reaches zero, i.e., $\lim_{t\rightarrow\infty}\|y(t)\|_{\B{Y}_t^{\mathrm{d}}}=0$. 
This goal should be reached quickly} while satisfying the combined constraints~\eqref{eq:obstacle_constraint_set} for all $t\in\B{N}$.
Furthermore, the approach should flexibly adapt to changes in the target $\B{Y}_t^{\mathrm{d}}$ during online operation.

\section{Proposed method}
\label{sec:method}
The basic idea of the proposed approach is illustrated in Figure~\ref{fig:illustrate_offset_stuck}. 
The target we want to reach can be far away and a possible path may be obstructed with obstacles. 
For computational reasons, we can only optimize motions over a short prediction horizon. Thus, we typically cannot directly optimize a trajectory that reaches the target.  
A key ingredient in the proposed formulation is the artificial reference, which is a setpoint jointly optimized by the MPC. The MPC optimizes a trajectory that reaches this reference, while an additional offset cost captures the difference between this artificial reference and the actual desired target. 
By suitably choosing this offset cost, the closed-loop system will converge to the desired target.

First, we first introduce collision-free references (Sec.~\ref{sec:method_steady_states}) and the proposed offset cost (Sec.~\ref{sec:method_offset}). Then, we  present the proposed MPC formulation (Sec.~\ref{sec:method_MPC}).

\subsection{Collision-free references}
\label{sec:method_steady_states}
The MPC optimizes an artificial reference
$y^{\mathrm{s}}\in\B{Y}$. 
To ensure reliable planning, this reference needs to correspond to a feasible steady-state $r^{\mathrm{s}}=(x^{\mathrm{s}},u^{\mathrm{s}})$\revised{, i.e., the superscript $\mathrm{s}$ indicates a steady state}. 
To this end, we introduce the (strictly) feasible  steady-state manifold
\begin{equation} \label{eq:steady_state}
    \B{S} := \{r^{\mathrm{s}} \mid r^{\mathrm{s}} = (x^{\mathrm{s}}, u^{\mathrm{s}}) \in \B{Z}_{\mathrm{r}}, x^{\mathrm{s}} = f(x^{\mathrm{s}}, u^{\mathrm{s}}) \},
\end{equation}
where $\B{Z}_{\mathrm{r}}\subseteq \operatorname{int}(\B{Z})$ is a compact set, which is chosen slightly smaller for technical reasons (cf.~\cite{krupa2024model}).
We denote the set of feasible steady-state \revised{positions} by
\begin{equation}
\B{S}_{\mathrm{y}}=\{y\in\B{Y}|~\exists (x,u)\in \B{S}, y=h(x)\}.
\label{eq:steady_output}    
\end{equation} 
This set does not yet account for the obstacles, and typically it is given by a simple box constraint characterizing the feasible domain in the Cartesian space $\B{Y}$. 

For the artificial reference $y^{\mathrm{s}}\in \B{S}_{\mathrm{y}}$, we derive sufficient conditions for the collision avoidance requirement~\eqref{eq:collision_avoidance} that only depend on the position $y=h(x)$ instead of the full geometry $\C{H}(x)$. 
Specifically, we compute the vehicular clearance $\delta_{\C{H}}>0$, which satisfies $\C{H}(x)\subseteq \{h(x)\}\oplus \B{B}_{\delta_{\C{H}}}$ for all $x\in\B{R}^n$, where $\B{B}_{\delta_{\C{H}}}$ is a ball of radius $\delta_{\C{H}}$. 
The radius  $\delta_{\C{H}}>0$ allows for an orientation independent characterization of the geometry.  
We define the inflated stationary obstacle  distance $\delta_{\mathrm{so}}:=\delta_{\mathrm{obst}}+\delta_{\C{H}}+\delta_{\epsilon}$.
The constant $\delta_{\epsilon}>0$ is a small extra buffer needed for technical reasons, while the factor $\delta_{\C{H}}$ ensures that we can directly pose the collision avoidance requirement now on the position $h(x)$ instead of the full geometry $\C{H}(x)$. 
Hence, we define the obstacle free \revised{position} space, including buffer, by 
\begin{equation}
\label{eq:steady_output_obstacle}    
\B{S}_{\mathrm{y},\B{O}}=\{y\in\B{S}_{\mathrm{y}}\mid \|y\|_{\B{O}_j}\geq \delta_{\mathrm{so}},~j\in\B{N}_{1:n_o}\}
\end{equation}
 and the corresponding set of steady-states as 
\begin{equation}
\label{eq:steady_state_obstacle}    
\B{S}_{\B{O}}=\{(x,u)\in \B{S}|~\|h(x)\|_{\B{O}_j}\geq \delta_{\mathrm{so}},~j\in\B{N}_{1:n_o}\}.
\end{equation} 
We define the stationary state and inputs achieving the target position $\B{Y}_t^{\mathrm{d}}$ as  
\begin{equation}
\label{eq:steady_state_optimal}
\B{Z}^{\mathrm{d}}_t=\{(x,u)\in\B{S}_{\B{O}}\mid h(x)\in\B{Y}_t^{\mathrm{d}}\}
\end{equation}
and by $\B{X}^{\mathrm{d}}_t\subseteq\B{R}^n$ we denote the projection on the state, i.e., the set of optimal steady-states. 

To summarize the set notation: 
The sets $\B{S},\B{S}_{\mathrm{y}},\B{S}_{\B{O}},\B{S}_{\mathrm{y},\B{O}}$ introduced in~\eqref{eq:steady_state},\eqref{eq:steady_output},\eqref{eq:steady_output_obstacle},\eqref{eq:steady_state_obstacle} correspond to the steady-state manifold, where the subscript $\B{O}$ indicates that also collision avoidance are considered and the subscript $\mathrm{y}$ highlight projection on the \revised{position} $y=h(x)$. 
For the state and input constraints $\B{Z}$, the subscript $\B{O}$ in~\eqref{eq:obstacle_constraint_set} also indicates inclusion of the collision avoidance constraint and the superscript $\mathrm{d}$ in~\eqref{eq:steady_state_optimal} indicates the optimal (\textbf{d}esired) set.
\subsection{Offset cost for non-convex environments}
\label{sec:method_offset}
We introduce an offset cost $T_{\B{Y}_t^{\mathrm{d}}}(y^{\mathrm{s}})$ to penalize the difference between the artificial setpoint $y^{\mathrm{s}}$ and the desired target $\B{Y}_t^{\mathrm{d}}$. 
Typically, the distance $\|y^{\mathrm{s}}\|_{\B{Y}_t^{\mathrm{d}}}$ is directly minimized, which can ensure convergence/stability of $\B{Y}_t^{\mathrm{d}}$ under suitable convexity conditions (cf.~\cite{limon2018nonlinear,krupa2024model}). 
However, as we show in experiments later (Sec.~\ref{sec:exp}), this approach is not applicable in environments cluttered with obstacles. 
To address this problem, we propose a new offset cost function, which is illustrated in Figure~\ref{fig:illustrate_offset_stuck}.

The offset cost is chosen as the length of the shortest path that connects the reference $y^{\mathrm{s}}$ to the target set $\B{Y}_t^{\mathrm{d}}$, while lying in the obstacle free \revised{position} space $\B{S}_{\mathrm{y},\B{O}}$~\eqref{eq:steady_output_obstacle}.
This approach is inspired by and refines the ideas in~\cite[Remark~3]{soloperto2022nonlinear}. 
We optimize over a piece-wise path given by a fixed number of $n_s\in\B{N}$ segments\revised{, which is represented by $y_\cdot\in\B{Y}^{n_{\mathrm{s}}}$}. 
The proposed offset cost is given by
\begin{subequations}    
\label{eq:offset_cost_segment}
\begin{align}
\label{eq:offset_cost_segment_cost}
T_{\B{Y}_t^{\mathrm{d}}}(y^{\mathrm{s}})=k_{\mathrm{M}}&\min_{\revised{y_{\cdot}}\in\B{Y}^{n_{\mathrm{s}}}}\sum_{j=0}^{n_s-1}\|y_{j+1}-y_{j}\|\\
\label{eq:offset_cost_segment_init}
\text{s.t. }& y_0=y^{\mathrm{s}},\\
\label{eq:offset_cost_segment_space_obstacle}
&\overline{y_jy_{j+1}}\in \B{S}_{\mathrm{y},\B{O}},~j\in\B{N}_{0:n_{\mathrm{s}}-1}\\
\label{eq:offset_cost_segment_end}
&y_{n_s}\in\B{Y}_t^{\mathrm{d}},
\end{align}
\end{subequations}
where $k_{\mathrm{M}}>0$ is a user chosen penalty. 
This offset cost optimizes over $n_{\mathrm{s}}$ segments that connect the artificial reference $y^{\mathrm{s}}$ to the target zone $\B{Y}_t^{\mathrm{d}}$. The offset cost is proportional to the length of the shortest $n_s$-segment path. 
The constraint~\eqref{eq:offset_cost_segment_space_obstacle} ensures that the segments lie in the obstacle-free \revised{position} space~\eqref{eq:steady_output_obstacle}.

\subsection{MPC formulation}
\label{sec:method_MPC}
\revised{Next, we introduce the MPC formulation. We use $x_{k|t}$ to denote the prediction of the state $k$ steps in the future starting from the measured state $x(t)$ at time $t$ and $u_{k|t}$ to denote the predicted (optimized) input. We also  denote the complete predicted sequence $x_{k|t}$,  $k\in\B{N}_{0:N}$ by $x_{\cdot|t}$ and analogously $u_{\cdot|t}$. }
Given a measured state $x(t)\in\B{R}^n$ and a desired target $\B{Y}_t^{\mathrm{d}}\subseteq\B{Y}$ at time $t\in\B{N}$, 
the MPC formulation is given by
\begin{subequations} \label{eq:mpc_opt}
\begin{align}
    \min_{u_{\cdot|t},x_{\cdot|t}, r_t^{\mathrm{s}}}
        \quad
    &\sum_{k=0}^{N-1}\ell(x_{k|t},u_{k|t},r_t^{\mathrm{s}} ) + T_{\B{Y}_t^{\mathrm{d}}}(y_t^{\mathrm{s}}),\\
        \text{s.t.} \quad
    &x_{k+1|t} = f(x_{k|t}, u_{k|t}), \quad  k \in \B{N}_{0:N-1}, \\
    \label{eq:mpc_opt_obstacle}
    &(x_{k|t}, u_{k|t}) \in \B{Z}_{\B{O}},\quad  k \in \B{N}_{0:N-1}, \\
    \label{eq:mpc_opt_steady_state}
    &r_t^{\mathrm{s}} =(x_t^{\mathrm{s}}, u_t^{\mathrm{s}}) \in \B{S}_{\B{O}},\quad y_t^{\mathrm{s}}=h(x_t^{\mathrm{s}}),\\
\label{eq:mpc_opt_initial}  &x_{0|t} = x(t),\\
\label{eq:mpc_opt_terminal}    &x_{N|t} = x^{\mathrm{s}}_t,
\end{align}
\end{subequations}
where $N\in\B{N}$ is the prediction horizon and $\ell$ is a positive definite stage cost. 
\revised{The MPC minimizes the distance between the predicted state and input trajectory and the artificial reference $r^{\mathrm{s}}$ using a positive definite stage cost $\ell(x,u,r^{\mathrm{s}})$~\cite{rawlings2017model,krupa2024model}.} 
The offset cost $T_{\B{Y}_t^{\mathrm{d}}}(y^{\mathrm{s}}_t)$ minimizes the difference of this artificial reference and the target $\B{Y}^{\mathrm{d}}_t$\revised{, i.e., the MPC also optimizes over the segments using the constraints and cost from~\eqref{eq:offset_cost_segment}}.  
For a fixed reference $r^{\mathrm{s}}_t$, Problem~\eqref{eq:mpc_opt} is a standard stabilizing MPC~\cite{rawlings2017model}, minimizing the distance to the reference over a finite-horizon, enforcing constraints~\eqref{eq:mpc_opt_obstacle}, and using a terminal equality constraint~\eqref{eq:mpc_opt_terminal}. 
Hence, Problem~\eqref{eq:mpc_opt} provides a unified formulation, combining a stabilizing MPC and a global planner. 
We assume that a unique minimizer to Problem~\eqref{eq:mpc_opt} exists, which is denoted by 
$x_{\cdot|t}^\star$, $u_{\cdot|t}^\star$, $r_t^{\mathrm{s},\star}$. 
At each time $t\in\B{N}$, we solve Problem~\eqref{eq:mpc_opt} and apply the first part of the optimal input sequence, i.e., $u(t)=u^\star_{0|t}$.  

\begin{remark}[Terminal constraint]
The proposed formulation uses a terminal equality constraint~\eqref{eq:mpc_opt_terminal} to simplify the exposition and the design. 
This can be naturally relaxed by designing a suitable terminal penalty~\cite{koehler2020nonlinearTAC} or by using a sufficiently long enough prediction horizon~\cite{kohler2024analysis}, see~\cite{limon2018nonlinear,soloperto2022nonlinear,krupa2024model}.
\end{remark}

 \section{Theoretical analysis}
\label{sec:theory}
We first detail the exact assumptions before deriving the theoretical guarantees. 
\subsection{Assumptions}
In the following, we detail and explain the assumptions on the cost, dynamics, and constraints.  

\begin{assumption}[Regularity]
\label{assump:regular}
The functions $f$, $h$, and $\ell$ are continuous \revised{with respect to their arguments.}\\ 
The sets $\B{Z},\B{Z}_{\mathrm{r}},\B{S},\B{S}_{\mathrm{y}}$ are compact. 
\end{assumption}    
\revised{Assumption~\ref{assump:regular} is a standard regularity condition on the involved sets and functions~\cite[Sec.~2.2]{rawlings2017model}}. 
\begin{assumption}[Regularity of steady-state manifold] \label{assump:unique}
There exists a constant $k_{\mathrm{Y}} > 0$ such that for any $r_1^{\mathrm{s}} = (x_1^{\mathrm{s}}, u_1^{\mathrm{s}}) \in \B{S}_{\B{O}}$ and $y_1^{\mathrm{s}}, y_2^{\mathrm{s}} \in \B{S}_{\mathrm{y},\B{O}}$, with $y_1^{\mathrm{s}} = h(x_1^{\mathrm{s}}, u_1^{\mathrm{s}})$, there exists some $r_2^{\mathrm{s}} = (x_2^{\mathrm{s}}, u_2^{\mathrm{s}}) \in \B{S}_{\B{O}}$ such that $h(x_2^{\mathrm{s}}, u_2^{\mathrm{s}}) = y_2^{\mathrm{s}}$ and
\begin{equation} \label{eq:assump:unique}
\| r_1^{\mathrm{s}} - r_2^{\mathrm{s}} \| \le k_{\mathrm{Y}} \| y_1^{\mathrm{s}} - y_2^{\mathrm{s}} \|.
\end{equation}
\end{assumption}
This condition is a relaxed version of~\cite[Asm.~1]{limon2018nonlinear}, \cite[Asm.~6]{krupa2024model}, which assumes that the \revised{position} $y^{\mathrm{s}}$ uniquely specifies a steady-state. 
Our relaxation is crucial as the orientation of a robot is not uniquely specified by the Cartesian coordinates $y$. For most mobile robot models, Condition~\eqref{eq:assump:unique} is trivially satisfied with $r^{\mathrm{s}}_2-r^{\mathrm{s}}_1=(y^{\mathrm{s}}_2-y^{\mathrm{s}}_1, \mathbf{0}_{n-p})$ and  $k_{\mathrm{Y}}=1$. 
\begin{assumption}[Path connectedness] \label{assump:link} 
The set $\B{S}_{\mathrm{y},\B{O}}$ is path connected with $n_s$-segments, i.e.,
for any two points $y_1, y_2 \in \B{S}_{\mathrm{y},\B{O}}$, there exists an $n_{\mathrm{s}}$-segment path given by waypoints $\{\mathscr{Y}_0,\dots \mathscr{Y}_{n_{\mathrm{s}}}\}\subseteq\B{S}_{\mathrm{y},\B{O}}$ with $\overline{\mathscr{Y}_j \mathscr{Y}_{j+1}}\subseteq\B{S}_{\mathrm{y},\B{O}}$, $j\in\B{N}_{0:n_{\mathrm{s}}-1}$, $\mathscr{Y}_0=y_1$, $\mathscr{Y}_{n_{\mathrm{s}}}=y_2$.  
\end{assumption}
Connectedness of the space $\B{S}_{\mathrm{y},\B{O}}$ is required to reach any feasible point, which can be assumed wlog, by otherwise restricting the analysis to the connected subset of $\B{S}_{\mathrm{y},\B{O}}$. 
Assumption~\ref{assump:link} also implies that the maximal required number of segments $n_s$ is utilized in the implementation in Problem~\eqref{eq:offset_cost_segment}. 
We relax this requirement in Section~\ref{sec:efficient_intermediate} to improve computational efficiency. 
\begin{assumption}[Feasible target]
\label{assump:feasible_target}    
For all $t\in\B{N}$,  $\B{Y}^{\mathrm{d}}_t\cap \B{S}_{\mathrm{y},\B{O}}\neq \emptyset$, i.e., 
the target set lies (at least partially) in the obstacle-free \revised{position} space.
\end{assumption}
\cref{assump:feasible_target,assump:link} ensure that it is possible to reach the target and the set of optimal steady-states $\B{Z}^{\mathrm{d}}_t$~\eqref{eq:steady_state_optimal} is non-empty. 
\begin{assumption}[Positive definite stage cost] \label{assump:stage_cost_bounds}
There exist $\underline{\alpha}_\ell,\bar{\alpha}_\ell\in\mathcal{K}_{\infty}$, such that for all $(x,u)\in\B{Z}$, $r^{\mathrm{s}}=(x^{\mathrm{s}},u^{\mathrm{s}})\in\B{S}$:
\begin{equation} \label{eq:assump:stage_cost_bounds}
\underline{\alpha}_\ell (\lVert x - x^{\mathrm{s}} \rVert) \leq \ell(x,u,r^{\mathrm{s}})\leq \overline{\alpha}_\ell(\|x-x^{\mathrm{s}}\|+\|u-u^{\mathrm{s}}\|).
\end{equation}
\end{assumption}
\revised{Assumption~\ref{assump:stage_cost_bounds} holds by choosing a  positive definite stage cost $\ell$, e.g., the standard quadratic cost $\ell(x,u,r^{\mathrm{s}})=\|x-x^{\mathrm{s}}\|_Q^2+\|u-u^{\mathrm{s}}\|_R^2$ with positive definite weighting matrices $Q,R$~\cite{rawlings2017model}.}
\begin{assumption}[Local $N$-step controllability] 
\label{assump:local_controllable}
There exist constants $\delta, k_{\mathrm{V}}>0$,  
such that for all feasible steady-states $(x_t^{\mathrm{s}},u_t^{\mathrm{s}})=r_t^{\mathrm{s}} \in\B{S}_{\B{O}}$ and any state $x(t)\in\B{R}^n:$ $\|x(t)-x_t^{\mathrm{s}}\|\leq \delta$, there exists a sequence $(x_{\cdot,t},u_{\cdot|t})\in\B{Z}_{\B{O}}^N$ satisfying the constraints in Problem~\eqref{eq:mpc_opt} and the following bound on the finite-horizon cost:
\begin{align}
\label{eq:local_upper_bound}
\sum_{k=0}^{N-1}\ell(x_{k|t},u_{k|t},r_t^{\mathrm{s}})\leq 
k_{\mathrm{V}}\|x(t)-x_t^{\mathrm{s}}\|^2.
\end{align}
\end{assumption}
Local feasibility follows from local $N$-step controllability, since $r^{\mathrm{s}}$ is subject to tighter constraints $\B{S}_{\B{O}}\subseteq \mathrm{int}(\B{Z}_{\B{O}})$ with $\B{Z}_{\mathrm{r}}\subseteq\mathrm{int}(\B{Z})$ and the extra buffer $\delta_\epsilon>0$ in the obstacle avoidance constraints $\B{S}_{\B{O}}$~\eqref{eq:steady_state_obstacle}. 
The local upper bound on the tracking cost is comparable to condition~\cite[Asm.~6c)]{krupa2024model}, and holds, e.g., if $\ell$ is locally quadratically bounded and the linearization is controllable (cf.~\cite[Prop.~4]{koehler2020nonlinearAutomatica}). 
For some non-holonomic robot dynamics, satisfaction of Assumption~\ref{assump:local_controllable} may require the utilization of a non-quadratic stage cost $\ell$.\footnote{%
For homogeneous system dynamics that are null controllable, Inequality~\eqref{eq:local_upper_bound} can be ensured using the corresponding homogeneous stage cost from~\cite{coron2020model}, see also~\cite{rosenfelder2023model} for application to common non-holonomic mobile robot dynamics. We use a standard quadratic stage cost in the experiments.}

\revised{To provide an efficient implementation later in Section~\ref{sec:efficient}, we also consider the following standard condition.  
\begin{assumption}[Convex stationary position space]
\label{assump:convex_output}
The set of steady-state positions $\B{S}_{\mathrm{y}}$ in~\eqref{eq:steady_output} is convex.
\end{assumption}
For wheeled robots or aerial drones, the dynamics are independent of the position $y$. Hence, the set $\B{S}_{\mathrm{y}}$ can be represented as a box that specifies the area where the robot should operate. 
Note that this condition only concerns the set of steady-state positions $\B{S}_{\mathrm{y}}$ and not to the collision-free region $\B{S}_{\mathrm{y},\B{O}}$, which is usually non-convex. In essence, this condition ensures that all non-convexity is confined to the obstacles $\B{O}$, which are handled separately. }

\revised{The following remark discusses when some of the considered assumptions may be violated in practice and what impact this has.} 
\revised{\begin{remark}[Practical considerations]\label{rk:assumptions}
A key assumption in our work is that the model~\eqref{eq:dynamics} accurately reflects the system behaviour and that the locations of the obstacles are known and static. While small inaccuracies can often be tolerated in practice, significant uncertainties will invalidate the following theoretical results and may lead to collisions or feasibility issues in practice. 
If any of Assumptions~\ref{assump:regular}--\ref{assump:convex_output} is violated, the system may fail to successfully converge to the desired setpoint and instead might get stuck. Assumption~\ref{assump:regular} might be violated for discontinuous dynamics, such as contacts in legged motion, which require significant modifications that are beyond the scope of this paper. For the considered applications in wheeled or aerial robots, Assumptions~\ref{assump:unique}, \ref{assump:link}, and \ref{assump:convex_output} are naturally satisfied. 
Assumptions~\ref{assump:stage_cost_bounds} and \ref{assump:feasible_target} are typically non-restrictive, since the stage cost $\ell$ can be freely designed and a target $\B{Y}_t^{\mathrm{d}}$ inside the obstacle region can simply be projected onto the feasible set $\B{S}_{\mathrm{y},\B{O}}$ before passing it to the controller. 
Assumption~\ref{assump:local_controllable} can be violated in practice, depending on the controllability of the dynamics and the stage cost $\ell$, in which case the system may only converge to a neighbourhood of the desired target instead of exactly reaching it; see \cite{muller2017quadratic}.
\end{remark}}
\subsection{Theoretical properties}
We first establish auxiliary results related to the proposed offset cost in Propositions~\ref{prop:offset_properties}--\ref{prop:art_reference_decrease} before deriving the closed-loop guarantees in Theorem~\ref{thm:main}. 

The following proposition summarizes key properties of the proposed offset cost~\eqref{eq:offset_cost_segment}. 
\begin{proposition}
\label{prop:offset_properties}
Let \cref{assump:regular,assump:unique,assump:link,assump:feasible_target} hold.
The offset cost $T_{\B{Y}_t^{\mathrm{d}}}(y^{\mathrm{s}})$ in~\eqref{eq:offset_cost_segment} is well defined, non-negative, and uniformly bounded for all feasible \revised{positions} $y^{\mathrm{s}}\in\B{S}_{\mathrm{y},\B{O}}$. \\
For any steady-state $(x^{\mathrm{s}},u^{\mathrm{s}})=r^{\mathrm{s}}\in\B{S}_{\B{O}}$, $y^{\mathrm{s}}=h(x^{\mathrm{s}})$ satisfies
\begin{align}
\label{eq:offset_upper_bound}
 \dfrac{k_{\mathrm{M}}}{k_\mathrm{Y}}\|r^{\mathrm{s}}\|_{\B{Z}_t^{\mathrm{d}}}\leq k_{\mathrm{M}}\|y^{\mathrm{s}}\|_{\B{Y}_t^{\mathrm{d}}}\leq T_{\B{Y}_t^{\mathrm{d}}}(y^{\mathrm{s}}). 
\end{align}
Furthermore, for any $\epsilon\in[0,1]$, there exists a steady-state $(\hat{x}^{\mathrm{s}},\hat{u}^{\mathrm{s}})=\hat{r}^{\mathrm{s}}\in\B{S}_{\B{O}}$, $\hat{y}^{\mathrm{s}}=h(\hat{x}^{\mathrm{s}})$, such that 
\begin{subequations}
\label{eq:offset_epsilon}    
\begin{align}
\label{eq:offset_epsilon_1}    
T_{\B{Y}_t^{\mathrm{d}}}(\hat{y}^{\mathrm{s}})\leq& (1-\epsilon)\cdot T_{\B{Y}_t^{\mathrm{d}}}(y^{\mathrm{s}}),\\
\label{eq:offset_epsilon_2}    
\|r^{\mathrm{s}}-\hat{r}^{\mathrm{s}}\|\leq & \epsilon\cdot k_{\mathrm{Y}} \cdot T_{\B{Y}_t^{\mathrm{d}}}(y^{\mathrm{s}}).
\end{align}
\end{subequations}
\end{proposition}
Condition~\eqref{eq:offset_upper_bound} ensures that a small offset cost $T_{\B{Y}^{\mathrm{d}}_t}$ also yields a small distance to the set of optimal states $\B{X}^{\mathrm{d}}_t$. Furthermore, Inequalities~\eqref{eq:offset_epsilon} ensure that the offset can be incrementally reduced. 
These conditions are comparable to~\cite[Assumption~3]{soloperto2022nonlinear}, which characterizes the desired properties of the offset cost,  independent of convexity. 

The following proposition establishes suitable properties of the artificial reference optimized in the MPC~\eqref{eq:mpc_opt}.
\begin{proposition}
\label{prop:art_reference_decrease}
Let \cref{assump:regular,assump:unique,assump:link,assump:stage_cost_bounds,assump:local_controllable,assump:feasible_target} hold.
There exists a constant $a>0$, such that if Problem~\eqref{eq:mpc_opt} is feasible at time $t\in\B{N}$, the optimal solution satisfies
\begin{align}
\label{eq:artificial_reference_distance}
 \|x(t)-x_t^{\mathrm{s},\star}\|\geq a\|r^{\mathrm{s},\star}_t\|_{\B{Z}_t^{\mathrm{d}}}.   
\end{align}
\end{proposition}
The derived bound~\eqref{eq:artificial_reference_distance} ensures that the optimal artificial reference $r^{\mathrm{s},\star}_t$ always pushes the system towards the optimal set $\B{Z}_t^{\mathrm{d}}$, which is crucial to ensure convergence. 
Similar inequalities are derived in related works for MPC with artificial references, see, e.g., \cite[Lemma~1]{kohler2024analysis}. 
The key difference is that these bounds typically rely on  a convex \revised{position} space $\B{S}_{\mathrm{y}}$. 
In contrast, the proposed offset cost~\eqref{eq:offset_cost_segment} is applicable to the non-convex space $\B{S}_{\mathrm{y},\B{O}}$, which is cluttered with obstacles.

The following theorem establishes the closed-loop properties of the proposed MPC scheme.
\begin{theorem}
\label{thm:main}
Let \cref{assump:regular,assump:unique,assump:link,assump:stage_cost_bounds,assump:local_controllable,assump:feasible_target} hold and suppose that Problem~\eqref{eq:mpc_opt} is initially feasible at $t=0$ with $x(0)$.
Then, Problem~\eqref{eq:mpc_opt} is feasible and the closed-loop
system satisfies the input, state, and collision avoidance constraints $(x(t),u(t))\in\B{Z}_{\B{O}}$ for all $t\in\B{N}$,
independent of possible changes in the target $\B{Y}_t^{\mathrm{d}}$. 
Suppose further that the target $\B{Y}_t^{\mathrm{d}}$ is constant. 
Then,  the set of optimal steady-states $\B{X}^{\mathrm{d}}_t$ is asymptotically\footnote{%
In case the stage cost $\ell$ is quadratic, Inequalities~\eqref{eq:Lyap_H_MPC} in the proof also imply exponential stability.  
} stable and $\lim_{t\rightarrow\infty}\|y(t)\|_{\B{Y}_t^{\mathrm{d}}}=0$. 
\end{theorem}
Problem~\eqref{eq:mpc_opt} decouples feasibility from the target $\B{Y}^{\mathrm{d}}_t$ by using artificial references $y^{\mathrm{s}}$. This ensures that we can ensure reliable operation (recursive feasibility, constraint satisfaction, collision avoidance), independent of online changes in the target $\B{Y}^{\mathrm{d}}_t$. 
Furthermore, Theorem~\ref{thm:main} ensures that we asymptotically stabilize the optimal set $\B{X}^{\mathrm{d}}_t$, which is the set of steady states that achieves $y(t)\in\B{Y}^{\mathrm{d}}_t$ while (strictly) satisfying the constraints (cf. \eqref{eq:steady_state_optimal}). 
Overall, these theoretical guarantees are comparable to existing results in MPC using artificial references~\cite{krupa2024model}; the main novelty lies in effectively addressing the non-convex environment $\B{Y}_{\B{O}}$ cluttered with obstacles, by using the offset cost in~\eqref{eq:offset_cost_segment}. 
Next, we focus on improving the formulation to allow for an efficient implementation.

\section{Efficient implementation}
\label{sec:efficient} 
In the following, we describe how Problem~\eqref{eq:mpc_opt} can be solved more efficiently.
First, we introduce an equivalent reformulation of the collision avoidance constraints (Sec.~\ref{sec:efficient_obstacle}). 
Then, we introduce the shortest path roadmap (Sec.~\ref{sec:efficient_roadmap}), which facilitates efficient warm-starting under changing targets $\B{Y}^{\mathrm{d}}_t$ using a global planner. 
Furthermore, we introduce intermediate targets to reduce the computational complexity of the offset cost (Sec.~\ref{sec:efficient_intermediate}). 
Lastly, we summarize the overall algorithm and establish its theoretical properties (Sec.~\ref{sec:efficient_alg_theory}). 

\subsection{Obstacle avoidance constraints}
\label{sec:efficient_obstacle} 
The obstacle avoidance constraints~\eqref{eq:offset_cost_segment_space_obstacle} and \eqref{eq:mpc_opt_obstacle} cannot be directly implemented in standard MPC solvers. 
Instead, we require (twice) continuously differentiable inequality constraints for an efficient implementation. 
To this end, we leverage the reformulation in~\cite{dietz2023efficient}, which introduces \emph{collision multipliers} as additional decision variables.  
\begin{lemma}(\cite[Prop.~1]{dietz2023efficient})
\label{lemma:obstacle_avoidance_dietz}
The following two statements are equivalent:
\begin{itemize}
    \item $\dist{\B{O}_i,\C{H}(x)}\geq \delta_{\mathrm{obst}}$
    \item there exist collision multipliers $\gamma_i:=[\mu^{\mathrm{r}}_i,\mu^{\mathrm{o}}_i,\xi_i^\top]^\top\in\B{R}^{p+2}$, such that:
    \begin{subequations}    
    \label{eq:collision_avoidance_lemma}
    \begin{align}
\label{eq:collision_avoidance_lemma_1}
   \mu^{\mathrm{r}}_i +\mu^{\mathrm{o}}_i+\frac{1}{4}\xi_i^\top \xi_i   + \delta_{\mathrm{obst}}^2\leq 0, \\
\label{eq:collision_avoidance_lemma_2}
    -\vertex{\C{H}(x)}^\top \xi_i - \mu^{\mathrm{r}}_i \mathbf{1}_{n_{\mathrm{vert},\C{H}}} \le 0, \\
\label{eq:collision_avoidance_lemma_3}
    \vertex{\B{O}_i}^\top \xi_i - \mu^{\mathrm{o}}_i \mathbf{1}_{n_{\mathrm{vert},\B{O}_i}} \le 0.
    \end{align}
    \end{subequations}
\end{itemize}
\end{lemma}
Inequalities~\eqref{eq:collision_avoidance_lemma_2} and \eqref{eq:collision_avoidance_lemma_3} correspond to two parallel half-space constraints, separating the obstacle and the vehicle geometry. 
Conditions~\eqref{eq:collision_avoidance_lemma} can be directly implemented in standard MPC solvers using the collision multipliers as additional decision variables. 
Note that the collision avoidance constraints~\eqref{eq:offset_cost_segment_space_obstacle} also represent collision avoidance between two polytopes and can be implemented analogous to Lemma~\ref{lemma:obstacle_avoidance_dietz} using $\vertex{\overline{y_j,y_{j+1}}}=[y_j,y_{j+1}]\in\B{R}^{p\times 2}$.

\subsection{Shortest road map \& warm-starts for changing targets}
\label{sec:efficient_roadmap}
\begin{figure}[t]
    \centering
    \includegraphics[width=\linewidth]{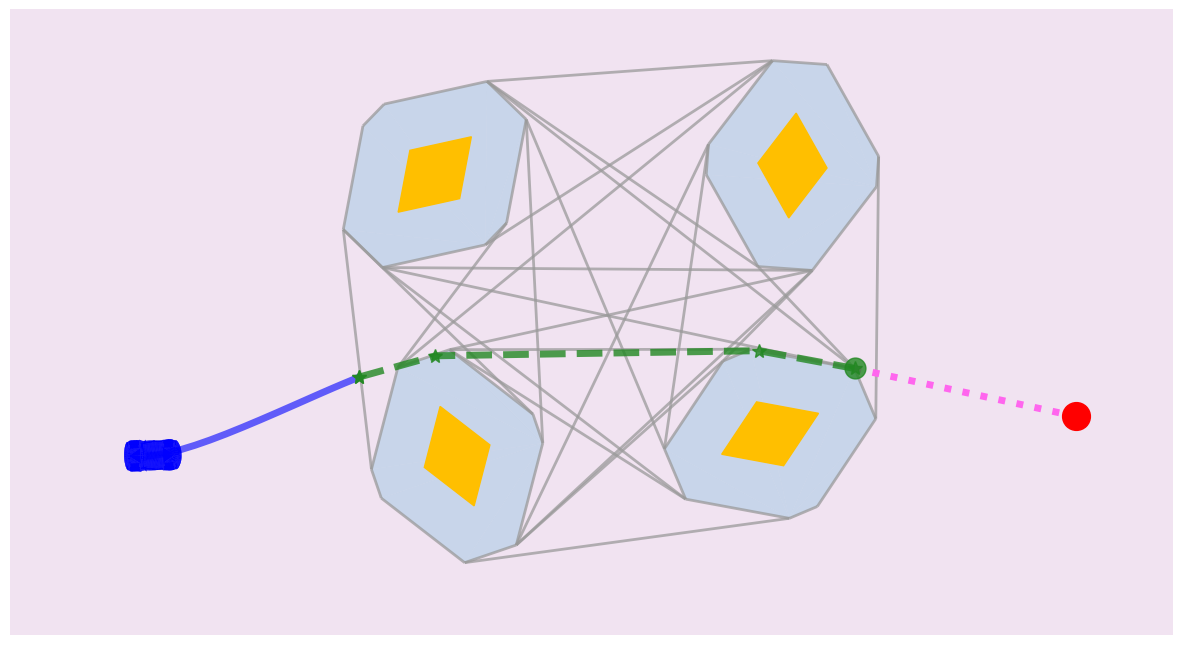}
\caption{Illustration of the shortest path road map with vehicle (blue), four obstacles (yellow), inflated obstacles (light blue region), and target (red). 
The roadmap is constructed from collision-free edges connecting vertices of inflated polytopic obstacles. 
The first and last segment connect the vehicle and the target to the graph and then remaining segments are the edges of the graph.}
\label{fig:shortestRoad}
\end{figure}

Recursive feasibility of Problem~\eqref{eq:mpc_opt} under online changing targets $\B{Y}^{\mathrm{d}}_t$ relies on the fact that the \revised{position} space is path connected (Asm.~\ref{assump:link}) and the target is feasible (Asm.~\ref{assump:feasible_target}). 
This ensures existence of a feasible solution to Problem~\eqref{eq:offset_cost_segment}. 
However, Problem~\eqref{eq:offset_cost_segment} is highly non-convex due to the obstacle constraints~\eqref{eq:offset_cost_segment_space_obstacle}, and thus standard (local) optimizers will not provide a feasible solution, unless initialized with a feasible solution. 
Hence, we use standard global path planning algorithms to determine the (globally) optimal path whenever the target $\B{Y}^{\mathrm{d}}_t$ changes. 
To this end, we create \revised{inflated} obstacles $\bar{\B{O}}_i$, which are compact polytopes satisfying 
\begin{align}
\label{eq:obstacle_inflated}    
\{y\in\B{Y} \mid \|y\|_{\B{O}_i}\leq \delta_{\mathrm{so}}+\delta_\epsilon\}\subseteq \bar{\B{O}}_i,
\end{align}
where $\delta_{\mathrm{so}}$ is the inflated stationary obstacle distance (Sec.~\ref{sec:method_steady_states}) and $\delta_\epsilon>0$ is a small offset for technical reasons. 
To account for the inflated obstacles $\bar{\B{O}}_i$, we consider the following stricter version of the collision-free \revised{position} space~\eqref{eq:steady_state_obstacle}:
\begin{align}
\label{eq:output_inflated_obstacle}
\bar{\B{S}}_{\mathrm{y},\B{O}}=\{y\in\B{S}_{\mathrm{y}}\mid y\cap \bar{\B{O}}_j=\emptyset,~j\in\B{N}_{1:n_\mathrm{o}}\}\subseteq \B{S}_{\mathrm{y},\B{O}}.    
\end{align}
Using the inflated obstacles $\bar{\B{O}}_i$, we create the shortest path road map~\cite[Sec.~6.2.4]{LaValle2006combinatorial}, which  is visualized in Figure~\ref{fig:shortestRoad}. 
The shortest path from a position $y$ to the target $\B{Y}^{\mathrm{d}}$ can be computed online using Dijkstra's algorithm, given the polytopic obstacles $\bar{\B{O}}_i$ and the convex \revised{position} space $\B{S}_{\mathrm{y}}$ (Asm.~\ref{assump:convex_output}).  
The resulting shortest path is characterized by a sequence of waypoints $\{\mathscr{Y}_0,\dots \mathscr{Y}_{n_{\mathscr{Y}}}\}\subseteq \B{S}_{\mathrm{y}}$ with 
$\mathscr{Y}_0=y$, $\mathscr{Y}_{n_{\mathscr{Y}}}\in\B{Y}^{\mathrm{d}}_t$ and $\overline{\mathscr{Y}_j\mathscr{Y}_{j+1}}\cap \overline{\B{O}}_i=\emptyset$, $j\in\B{N}_{1:n_{\mathscr{Y}}-1}$, $\overline{\mathscr{Y}_0\mathscr{Y}_{1}}\subseteq\B{S}_{\mathrm{y},\B{O}}$. 

During online operation, we need to compute shortest path starting from some position $y\in\B{S}_{\mathrm{y},\B{O}}$, which may not necessarily lie in $\bar{\B{S}}_{\mathrm{y},\B{O}}$. Thus, for the first link, $\overline{\mathscr{Y}_0 \mathscr{Y}_1}$, we only impose the more relaxed constraints $\B{S}_{\mathrm{y},\B{O}}$. This can be implemented naturally considering how the graph is constructed (cf. Fig.~\ref{fig:shortestRoad}).

This shortest path is used to initialize the decision variables $y_j$, $j\in\B{N}_{0:n_{\mathrm{s}}}$ for determining the offset cost $T_{\B{Y}^{\mathrm{d}}_t}$ in the MPC, whenever the target $\B{Y}^{\mathrm{d}}_t$ changes. 
Since the obstacle avoidance constraints~\eqref{eq:obstacle_constraint_set} are implemented using additional decision variables $\gamma$, we equally supply a feasible warm-start, which can be computed based on the geometric intuition of Lemma~\ref{lemma:obstacle_avoidance_dietz}.

\revised{\begin{remark}[Alternative global planners]
\label{rk:global_planner}   
We use Dijkstra's algorithm to compute the shortest path to a new target $\B{Y}^{\mathrm{d}}$ online due to its simplicity and the fact that it directly yields segments that can be used to initialize the MPC solver. 
However, sampling-based methods~\cite{orthey2023sampling} often exhibit better scalability for cluttered environments. 
The proposed framework can also be applied with such alternative global planners without any modifications to the MPC formulation. The only requirement is that the planner generates a path that stays in the feasible set $\bar{\B{S}}_{\mathrm{y},\B{O}}$. 
\end{remark}}

\begin{figure}[th]
    \centering
\includegraphics[width=\linewidth]{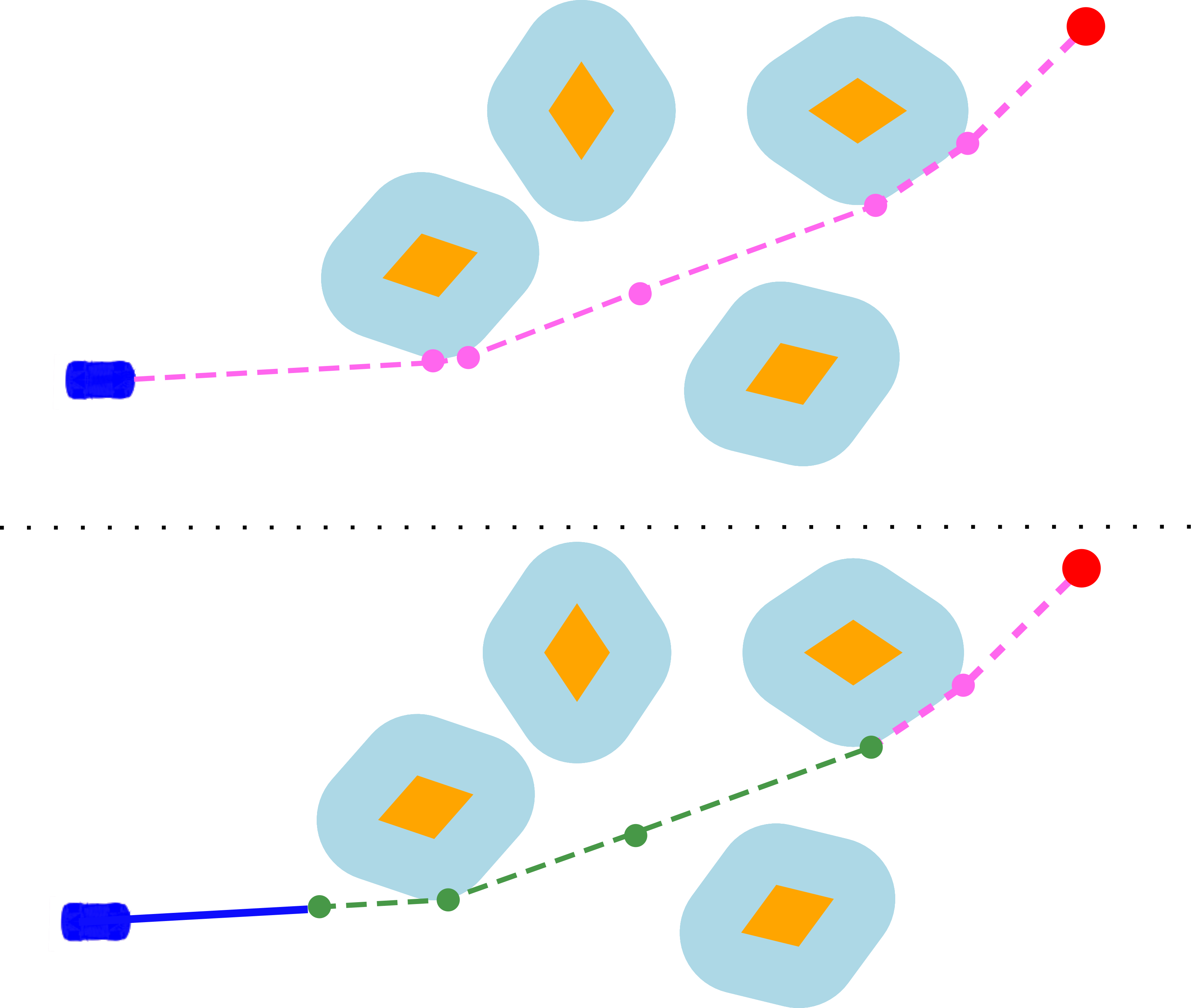}            \caption{Illustration of the intermediate target with obstacles (yellow), inflated obstacles (light blue), and target (red circle). 
    Top: shortest path calculated from the start to the target. 
    Bottom: The MPC optimizes over the predicted trajectory (blue, solid) and $n_\nu=3$ segments (green, dashed). 
    The segments end at the pre-computed shortest path (purple), which is incremented whenever possible.
    Note that the optimized segments differ from the pre-computed shortest path.}
    \label{fig:intermediate_target}
\end{figure}
\subsection{Intermediate targets}
\label{sec:efficient_intermediate}
The offset cost in Problem~\eqref{eq:offset_cost_segment} requires optimizing over $n_s$-segments, which can be computationally prohibitive for complex environments with many obstacles. 
To reduce the computational demand, we optimize only over $n_\nu\in \B{N}$ segments, where $n_\nu\in\B{N}$ is a user chosen constant satisfying $n_\nu\in[2,n_{\mathrm{s}}]$. 
\revised{This decreases the number of decision variables and constraints compared to Problem~\eqref{eq:mpc_opt}, which is crucial for fast real-time optimization.}\color{black}{ }
Instead of directly reaching the target $\B{Y}^{\mathrm{d}}$, we use an intermediate target $\hat{y}^{\mathrm{d}}\in\B{S}_{\mathrm{y},\B{O}}$. 
Figure~\ref{fig:intermediate_target} illustrates how the intermediate target, the optimized segments, and the shortest path are connected.

 The intermediate target is chosen as one of the waypoints $\mathscr{Y}_j$ of the path computed using the shortest road map (Section~\ref{sec:efficient_roadmap}). 
The selection and update of the intermediate target  $\hat{y}^{\mathrm{d}}$ needs to ensure that the MPC optimization problem remains feasible and convergence guarantees from Theorem~\ref{thm:main} remain valid. 
This is accomplished by incrementing the intermediate target $\hat{y}^{\mathrm{d}}$ along the shortest path waypoints $\mathscr{Y}$ using Algorithm~\ref{alg:condensing}. 
This algorithm uses $y_{j}$, the path segments optimized by the MPC, and $\mathscr{Y}_j$, the waypoints of the shortest path that have not yet been utilized. 
\revised{In order to increment the waypoint, the algorithm checks if we can skip one of the optimized points $y_{j+1}$ by directly connecting $y_j$ and $y_{j+2}$ with a straight line  while respecting the inflated obstacle avoidance constraints. 
If this is the case, the point $y_{j+1}$ is removed and we add the next waypoint from the shortest path $\mathscr{Y}$ to the optimized points, which becomes the new intermediate target $\hat{y}^{\mathrm{d}}$.} 
By construction, the algorithm also yields a sequence of waypoints $y_{\cdot}$ that are feasible for the offset cost in the MPC using the new intermediate target $\hat{y}^{\mathrm{d}}$, thus ensuring recursive feasibility. 

\begin{algorithm}[ht]
\caption{Increment intermediate target $\hat{y}^{\mathrm{d}}$}
\begin{algorithmic}
\REQUIRE Waypoints from shortest path 
    $\mathscr{Y}_j$, $j\in\B{N}_{0:n_{\mathscr{Y}}}$ and optimized segments $y_{j}$, $j\in\B{N}_{0:n_{\nu}}$. 
\STATE $j\leftarrow 0$. \COMMENT{Initialize loop.}
\WHILE{$j \leq n_{\nu}-2$ \& $n_{\mathscr{Y}}>0$ }
\STATE{\COMMENT{Check if waypoint can be skipped.}}
\IF{$\dist{\overline{y_{j} y_{j+2}},\B{O}_i}\geq \delta_{\mathrm{so}}$ $\forall i\in\B{N}_{1:n_{\mathrm{o}}}$ }
\STATE $y_{\cdot}\leftarrow \{y_{0:j},y_{j+2:n_\nu},\mathscr{Y}_0\}$. \\\COMMENT{Skip waypoint and append from shortest path.}
\STATE $\mathscr{Y}\leftarrow \{\mathscr{Y}_1,\dots \mathscr{Y}_{n_{\mathscr{Y}}}\}$, $n_{\mathscr{Y}}\leftarrow n_{\mathscr{Y}}-1$. \\\COMMENT{Remove waypoint from shortest path.}
                \ELSE
                \STATE $j\leftarrow j+1$.
            \ENDIF
\ENDWHILE
\STATE  $\hat{y}^{\mathrm{d}}\leftarrow y_{n_\nu}$.
\COMMENT{Set new intermediate target.}
\end{algorithmic}
\label{alg:condensing}
\end{algorithm}

Recall that the shortest road map is constructed using \revised{inflated} obstacles $\bar{\B{O}}_i$~\eqref{eq:obstacle_inflated}, which enforce a collision avoidance distance strictly larger than $\delta_{\mathrm{so}}$. 
This ensures that Algorithm~\ref{alg:condensing} successfully increments the intermediate target whenever the distance between optimized targets $y_{j}$ is sufficiently small. 
This property will be crucial for the convergence analysis in Theorem~\ref{thm:MPC_eff}.

 \subsection{Overall algorithm \& theoretical properties}
\label{sec:efficient_alg_theory}
In the following, we summarize the overall algorithm, which builds on the methodology in Section~\ref{sec:method}, but also uses the collision avoidance reformulation (Sec.~\ref{sec:efficient_obstacle}), the shortest road map (Sec.~\ref{sec:efficient_roadmap}), and the intermediate target (Sec.~\ref{sec:efficient_intermediate}) to enhance computational efficiency.

At each time $t$, we solve the following optimization problem using the current state $x(t)$ and the intermediate target $\hat{y}^{\mathrm{d}}_t$:
{
\begin{subequations} \label{eq:mpc_intermediate}
\begin{align}
    &\mathrm{minimize}_{u_{\cdot|t},x_{\cdot|t}, r_t^{\mathrm{s}},\bar{y}_{\cdot|t},\gamma_{\cdot|t},\bar{\gamma}_{\cdot|t}}\nonumber\\
    &\sum_{k=0}^{N-1}\ell(x_{k|t},u_{k|t},r_t^{\mathrm{s}} ) + k_{\mathrm{M}}\sum_{j=0}^{n_{\nu}-1}\|\bar{y}_{j+1|t}-\bar{y}_{j|t}\|   \\
        \text{s.t.} \quad
    &x_{k+1|t} = f(x_{k|t}, u_{k|t}), \quad  (x_{k|t}, u_{k|t}) \in \B{Z}, \\
\label{eq:mpc_intermediate_obstacle_x_start}
    &\mu_{k,i|t}^{\mathrm{r}}+\mu_{k,i|t}^{\mathrm{o}}+\dfrac{1}{4}\xi_{k,i|t}^\top \xi_{k,i|t}+\delta_{\mathrm{obst}}^2\leq 0,\\
    &-\vertex{\C{H}(x_{k|t})}^\top \xi_{k,i|t} - \mu^{\mathrm{r}}_{k,i|t} \mathbf{1}_{n_{\mathrm{vert},\C{H}}} \le 0 ,\\
\label{eq:mpc_intermediate_obstacle_x_end}
    &\vertex{\B{O}_i}^\top \xi_{k,i|t} - \mu^{\mathrm{o}}_{k,i|t} \mathbf{1}_{n_{\mathrm{vert},\B{O}_i}} \le 0,\\
    &k\in\B{N}_{0:N-1},\quad i\in\B{N}_{1:n_{\mathrm{o}}},\nonumber\\
    &f(x_t^{\mathrm{s}},u_t^{\mathrm{s}})=x_t^{\mathrm{s}},\quad r^{\mathrm{s}}_t=(x^{\mathrm{s}}_t,u_t^{\mathrm{s}})\in\B{Z}_{\mathrm{r}},\\
&x_{0|t} = x(t), \quad x_{N|t} = x^{\mathrm{s}}_t,\\ 
\label{eq:mpc_intermediate_offset_terminal}
& \bar{y}_{0|t}=y^{\mathrm{s}}_t=h(x^{\mathrm{s}}_t),\quad \bar{y}_{n_\nu|t}=\hat{y}^{\mathrm{d}}_t,\\
\label{eq:mpc_intermediate_obstacle_offset_start}
 & \bar{\mu}^{\mathrm{r}}_{j,i|t} +\bar{\mu}^{\mathrm{o}}_{j,i|t}+\frac{1}{4}\bar{\xi}_{j,i|t}^\top \bar{\xi}_{j,i|t}   + \delta_{\mathrm{so}}^2\leq 0, \\
&  -[\bar{y}_{j|t},\bar{y}_{j+1|t}]^\top \bar{\xi}_{j,i|t} - \bar{\mu}^{\mathrm{r}}_{j,i|t} \mathbf{1}_{2} \le 0,\\
\label{eq:mpc_intermediate_obstacle_offset_end}
& \vertex{\B{O}_i}^\top \bar{\xi}_{j,i|t} - \bar{\mu}^{\mathrm{o}}_{j,i|t} \mathbf{1}_{n_{\mathrm{vert},\B{O}_i}} \le 0, \\
& \bar{y}_{j|t}\in \B{S}_{\mathrm{y}},\\
&j\in\B{N}_{0:n_\nu-1},\quad i\in\B{N}_{1:n_{\mathrm{o}}}\nonumber. 
\end{align}
\end{subequations}}
Problem~\eqref{eq:mpc_intermediate} efficiently implements the obstacle avoidance constraints~\eqref{eq:offset_cost_segment_space_obstacle},  \eqref{eq:mpc_opt_obstacle},  \eqref{eq:mpc_opt_steady_state}  in~\eqref{eq:mpc_intermediate_obstacle_x_start}--\eqref{eq:mpc_intermediate_obstacle_x_end}
 and \eqref{eq:mpc_intermediate_obstacle_offset_start}--\eqref{eq:mpc_intermediate_obstacle_offset_end} 
using the reformulation  from Lemma~\ref{lemma:obstacle_avoidance_dietz} and convexity of $\B{S}_{\mathrm{y}}$ (\cref{assump:convex_output}).  
Compared to Problem~\eqref{eq:mpc_opt}, the offset cost~\eqref{eq:offset_cost_segment} is also replaced by a shorter $n_{\nu}$-segment offset cost that ends at the intermediate target $\hat{y}^{\mathrm{d}}_t$.

Algorithm~\ref{alg:overall} summarizes the closed-loop operation. 
Compared to the method presented in Section~\ref{sec:method_MPC}, the primary difference is the update of the intermediate target $\hat{y}^{\mathrm{d}}_t$ using Algorithm~\ref{alg:condensing} and the utilization of the waypoints $\mathscr{Y}(t)$ from the shortest path.
These differences significantly reduce the computational demand.

The following theorem shows that the theoretical guarantees  from Theorem~\ref{thm:main} (safety \& convergence) remain valid.
\begin{theorem}  
\label{thm:MPC_eff}
Let \cref{assump:regular,assump:unique,assump:stage_cost_bounds,assump:local_controllable,assump:convex_output} hold. Suppose that  $\bar{\B{S}}_{\mathrm{y},\B{O}}$ is path connected and $\B{Y}_t^{\mathrm{d}}\cap \bar{\B{S}}_{\mathrm{y},\B{O}}\neq \emptyset$, i.e., \cref{assump:link,assump:feasible_target} hold using the smaller set $\bar{\B{S}}_{\mathrm{y},\B{O}}\subseteq\operatorname{int}(\B{S}_{\mathrm{y},\B{O}})$~\eqref{eq:output_inflated_obstacle}. 
Suppose further that the system is initialized at a feasible steady-state (cf. Alg.~\ref{alg:overall}). 
Then, the closed-loop system resulting from Algorithm~\ref{alg:overall} satisfies: \\
I) (Recursive feasibility) All the optimization problems and algorithms in
Algorithm~\ref{alg:overall} are feasible for all $t\in\B{N}$.\\
II) (Constraint satisfaction) The resulting closed-loop trajectory satisfies the state and input constraints~\eqref{eq:constraints} and ensures collision avoidance~\eqref{eq:collision_avoidance}, for all $t\in\B{N}$.\\
III) If the target $\B{Y}_t^{\mathrm{d}}$ is constant, then we asymptotically converge\footnote{%
In contrast to Theorem~\ref{thm:main}, we do not establish asymptotic stability of the set $\B{X}^{\mathrm{d}}_t$, since the control input $u(t)$, is not only a function of the state $x(t)$, but also the intermediate target $\hat{y}_t^{\mathrm{d}}$.}
to $\B{Y}^{\mathrm{d}}_t$, i.e., 
$\lim_{t\rightarrow\infty}\|y(t)\|_{\B{Y}_t^{\mathrm{d}}}=0$.
\end{theorem}
Theorem~\ref{thm:MPC_eff} guarantees that Algorithm~\ref{alg:overall} ensures safety (constraint satisfaction and collision avoidance) as well as task completion (converges to the target). 
\begin{algorithm}[th]
\caption{Closed-loop operation}
\begin{algorithmic}
\STATE Start at feasible steady-state $x(0)$ with $y(0)\in\bar{\B{S}}_{\mathrm{y},\B{O}}$.
\STATE Initialize $\bar{y}_{\cdot|t-1}^\star=y^{\mathrm{s},\star}_{-1}=\hat{y}^\mathrm{d}_{-1}=y(0)$.
\FOR{$t=0,1,\dots$}
         \IF {$t = 0$ or $\B{Y}^{\mathrm{d}}_t\neq \B{Y}^{\mathrm{d}}_{t-1}$}
        \STATE $\mathscr{Y}(t)$ $\leftarrow$ shortest path from $y^{\mathrm{s},\star}_{t-1}$ to $\B{Y}^{\mathrm{d}}_t$.\\
        \COMMENT{Use shortest road map (Sec.~\ref{sec:efficient_roadmap}) to compute path.}
        \STATE $\bar{y}^\star_{j|t-1}\leftarrow \mathscr{Y}_j(t)$, $j\in\B{N}_{0:n_{\nu}}$,  
         $\mathscr{Y}(t)\leftarrow\{\mathscr{Y}_{n_\nu+1}(t),\dots,\mathscr{Y}_{n_{\mathscr{Y}(t)}}(t)\}$, $n_{\mathscr{Y}(t)}\leftarrow n_{\mathscr{Y}(t)}-n_\nu$,  $\hat{y}^{\mathrm{d}}_t\leftarrow \bar{y}_{n_\nu|t-1}^\star$. \COMMENT{Initialize new paths.}
        \ENDIF
        \IF{$\hat{y}^\mathrm{d}_t\notin\B{Y}^{\mathrm{d}}_t$}
        \STATE $\mathscr{Y}(t)$, $\hat{y}^{\mathrm{d}}_t$, $\bar{y}_{\cdot|t} \leftarrow$ Run Algorithm~\ref{alg:condensing} using $y\leftarrow \bar{y}^\star_{\cdot|t-1}$, $\mathscr{Y}\leftarrow \mathscr{Y}(t-1)$. \COMMENT{Increment intermediate target.}
        \ENDIF 
        \STATE $u^\star_{\cdot|t},\bar{y}_{\cdot|t}^\star$ $\leftarrow$ Solve Problem~\eqref{eq:mpc_intermediate}.
        \STATE Apply optimal input $u(t)=u^\star_{0|t}$.
\ENDFOR
    \end{algorithmic}
    \label{alg:overall}
\end{algorithm}

\section{Experimental results}
\label{sec:exp}
In this section, we demonstrate the performance of our controller in cluttered environments using a miniature car in simulation and hardware experiments. First, we provide details on the software and implementation. 
Then we provide simulation results, comparing the proposed controller to existing MPC formulations in environments with randomly placed obstacles.
Then, we show hardware experiments for the proposed method.
\subsection{Setup and implementation details}

The experiments are performed with a miniature RC car, scaled at 1:28, implemented in
CRS~\cite{carron2023CRS} using ROS~\cite{ros}.
\subsubsection*{Mobile robot dynamics}
The system is modelled using a kinematic bicycle model
\begin{align*}
\dot{x}
=&
\begin{bmatrix}
    v \cos(\theta + \beta) \\
    v \sin(\theta + \beta) \\
    v \sin(\beta) / l_{\mathrm{r}} \\
    (-v + a \cdot T)/\tau \\
    \Delta T \\
    \Delta \omega
\end{bmatrix}, \quad 
 x =\begin{bmatrix}
p_{\mathrm{x}}\\ p_{\mathrm{y}}\\ \theta\\ v\\ T\\ \omega
\end{bmatrix}\in\B{R}^6,\\
u =&\begin{bmatrix}
\Delta T\\ \Delta \omega    
\end{bmatrix}\in\B{R}^2,\quad 
y=\begin{bmatrix}
    p_{\mathrm{x}}\\p_{\mathrm{y}}
\end{bmatrix}\in\B{R}^2,\\
\beta=& \arctan(\tan (\omega) \cdot l_{\mathrm{r}} / (l_{\mathrm{f}} + l_{\mathrm{r}})),
\end{align*}
with Cartesian position $p_{\mathrm{x}},p_{\mathrm{y}}$, angle $\theta$, velocity $v$, torque $T$, steering angle $\omega$, change in torque $\Delta T$, change in steering $\Delta\omega$, and slip angle
 $\beta$. 
Model parameters are 
 $a = 5.03$, rear wheel distance $l_{\mathrm{r}} = \SI{5.17}{\cm}$, front wheel distance $l_{\mathrm{f}} = \SI{4.66}{\cm}$, and time constant $\tau = \SI{0.8}{\s}$.
The dynamics are integrated using 
an explicit Runge-Kutta 4th-order discretization. 

The car geometry is described by a rectangle of dimensions $\SI{12.8}{\cm} \times \SI{7.1}{\cm}$, rotated by $\theta$. 
All obstacles $\B{O}_i$ have a diamond shape (rhombus) with diagonal lengths $\SI{23.5}{\cm}$ and $\SI{15.5}{\cm}$. 
The vehicle radius is $\delta_{\C{H}}=\SI{7.3}{\cm}$, the obstacle distance $\delta_{\mathrm{obst}}=\SI{3.0}{\cm}$, and the extra buffer is $\delta_{\epsilon}=\SI{1.0}{\cm}$
\subsubsection*{Software}
The MPC problem~\eqref{eq:mpc_opt} is solved with Acados~\cite{verschueren2022acados} using real-time iterations (\verb|SQP_RTI|) and HPIPM~\cite{frison2020hpipm}. The segments $\bar{y}_{j|t}$ for the offset cost are implemented as a second phase in a multiphase optimal control problem~\cite{frey2025multi}. In particular, the second phase consists of $n_\nu$-stages with  integrator dynamics $\bar{y}_{j+1}=\bar{y}_j+\Delta \bar{y}_j$ and initial condition specified through $\bar{y}_0=h(x_N)$. 
The shortest path roadmap (Sec.~\ref{sec:efficient_roadmap}) is implemented using CGAL~\cite{cgal} for computational geometry and Boost Graph Library~\cite{siek2001boost} for graph operations. In CGAL, we used the Axis-Aligned Bounding Box (AABB) tree as a spatial data structure for efficient obstacle look-up and collision detection, implemented  with an inexact floating-point kernel. 
The MPC has a prediction horizon of $N=20$ and $n_{\nu}=3$ segments are optimized for the offset cost. 
All controllers were implemented with \revised{\SI{20}{\hertz}}. \\
We run both simulation and hardware experiments on a computer equipped with an Intel Core i9-13900KS CPU (base frequency 3.20 GHz) and 48 GB of RAM.
The complete implementation is available open source:
\begin{center}
{\small \url{https://github.com/IntelligentControlSystems/ClutteredEnvironment}}
\end{center}

\subsection{Simulation results}
\begin{figure}[t]
    \centering
        \includegraphics[width=0.485\textwidth]{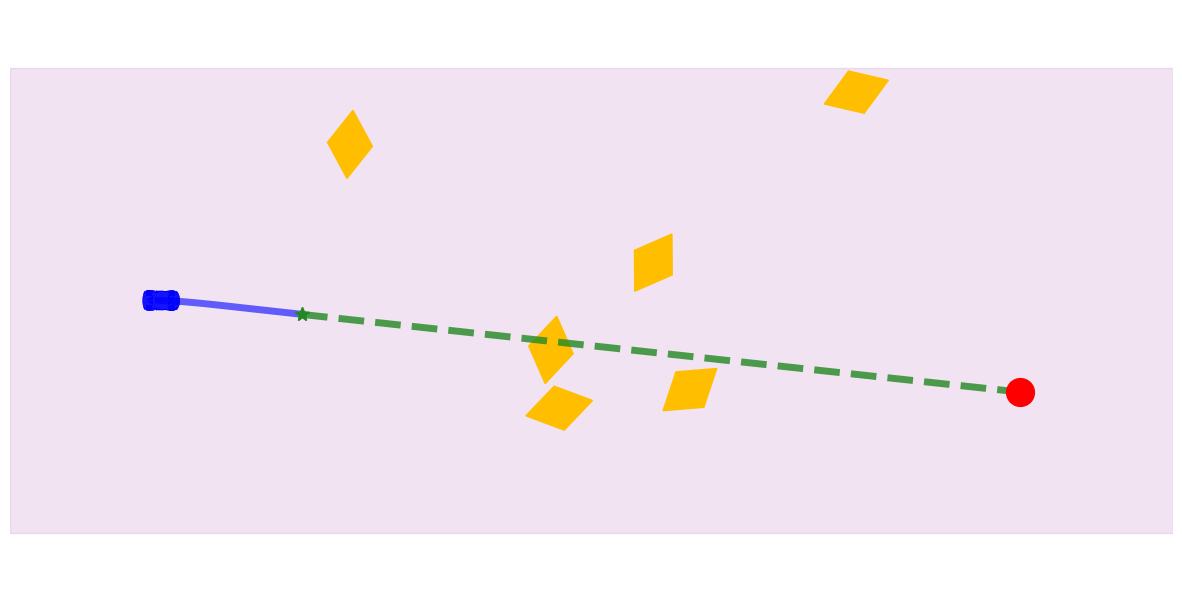}\\
        \includegraphics[width=0.485\textwidth]{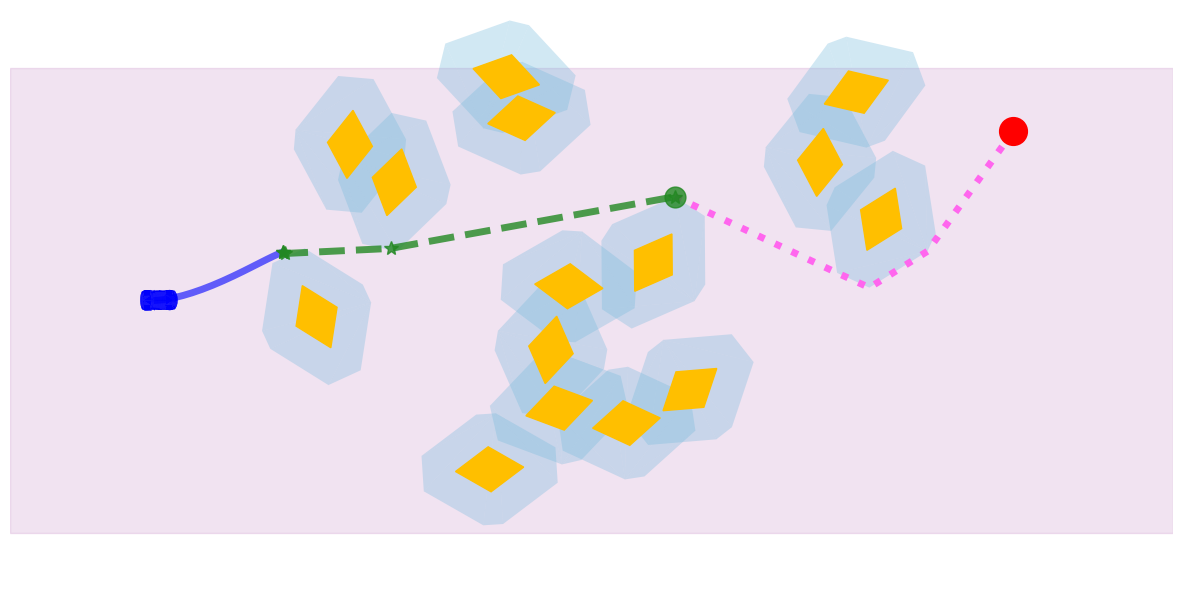}
    \caption{Visualization of exemplary randomly generated environments.  Car and predicted trajectory (blue, solid), target (red circle), obstacles (yellow), inflated obstacles (light blue). Top: Sparse environment with $n_{\mathrm{o}}=6$ obstacles and MPC with the $L_2$-norm offset cost (green, dashed). Bottom: Dense environment with $n_{\mathrm{o}}=15$ obstacles and MPC with proposed  segment-based offset cost, consisting of the optimized segmented (green, dashed), intermediate target \revised{$\hat{y}^{\mathrm{d}}_t$} (large green circle), and the remaining shortest path (purple, dotted). \revised{Since the $L_2$ formulation considers a straight line, the car would get stuck in front of the obstacle; while the proposed approach directly uses a collision-free path for navigation.} }
    \label{fig:environments}
\end{figure}

\begin{figure}[t]
    \centering
        \includegraphics[width=0.485\textwidth]{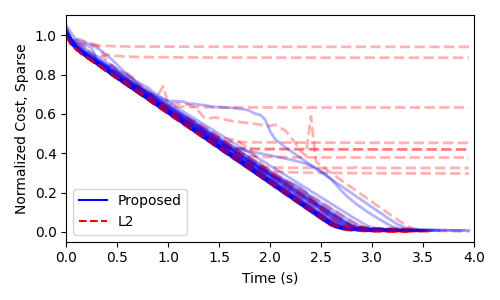}\\
        \includegraphics[width=0.485\textwidth]{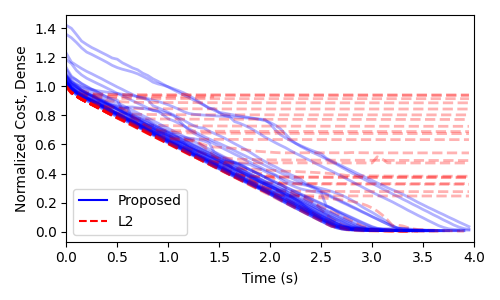}
    \caption{Closed-loop convergence of the proposed scheme with the proposed segment-based offset cost (blue) and the simple $L_2$-norm offset cost (red, dashed) in $30$  sparse environments (top) and $30$  dense environments (bottom). Cost includes the value function of the Problem~\eqref{eq:mpc_intermediate} and the length of the remaining shortest path, see Equation~\eqref{eq:augmented_cost_including_shortest_path} in the appendix. The cost is normalized  to the initial cost of the MPC with the simple $L_2$ offset cost and provides an estimate for the distance to the target.}
    \label{fig:sim_comparion}
\end{figure}

We compare two MPC schemes:
\begin{enumerate}
    \item \emph{Proposed}: The proposed approach, which jointly optimizes a $3$-segment path for the offset cost.
    \item \emph{$L_2$}: A standard MPC with artificial reference~\cite{limon2018nonlinear,krupa2024model}, where the offset cost is simply the euclidean distance to the target, i.e., $T(y^{\mathrm{s}})=k_{\mathrm{M}}\|y^{\mathrm{s}}-y^{\mathrm{d}}\|$.
\end{enumerate}
We evaluate both controllers for randomly generated environments. We distinguish \emph{sparse} environments with $n_{\mathrm{o}}=6$ obstacles and \emph{dense} environments with $n_{\mathrm{o}}=15$ obstacles. 
The obstacles are generated using random positions and orientations subject to the following constraints:
\begin{itemize}
\item Their centres lie within a specified obstacle region. 
\item The obstacles do not overlap.
\item There exists a path from the left side of the obstacle region to the right that remains a distance of $\SI{15}{\cm}$ away from the obstacles (cf. Assumption~\ref{assump:link}).
\end{itemize}
The initial position $y(0)$ and target point $\B{Y}^{\mathrm{d}}$ are randomly generated on opposite sides of the obstacle region, and the target is static. 
The setup is visualized in Figure~\ref{fig:environments}.
We compare the proposed MPC - using the segment-based offset cost - to a tracking MPC using a simple $L_2$-norm offset cost in $30$ sparse and $30$ densely cluttered environments. 
The proposed MPC was able to reach the target within $\SI{4}{\s}$ for $100\%$ of the considered environments. 
Thus, the proposed approach provides reliability (100$\%$ success rate) and high performance (fast execution). 
In contrast, the simple $L_2$-based MPC terminated successfully in $73\%$ of the sparse environments and only $27\%$ of the dense environments. In all other cases, the car got stuck at an obstacle and could not  progress further without \revised{violating the obstacle avoidance constraints}. 
The results confirm that our controller can reliably navigate through cluttered non-convex environments, while a standard $L_2$-norm offset cost often fails. 
Figure~\ref{fig:sim_comparion} shows the convergence behaviour for each trial. 
For the $L_2$-MPC, the cost is (mostly\footnote{%
While the standard candidate solution ensures that the optimal cost is non-increasing, this property may be lost when using a multiple shooting formulation and a real-time iteration, see~\cite{numerow2024inherently} for required modifications.}) monotonically decreasing, however, it often does not converge to zero. 
This indicates that the car converged to a steady-state close to an obstacle and cannot further progress towards the target, similar to the illustration in Figure~\ref{fig:illustrate_offset_stuck}. 
If the $L_2$-offset cost were replaced by a simple potential field (cf.~\cite{rasekhipour2016potential}), the MPC would similarly get stuck in local minima, as it lacks a global planner~\cite{khatib1986real}. In contrast, the cost of the proposed formulation decreases strictly until the target is reached, as guaranteed by Theorem~\ref{thm:MPC_eff}. 

\revised{Table~\ref{tab:sim_computation} lists the worst-case computation times measured across all time steps and all conducted simulation experiments. }  
For the proposed MPC, the \emph{Per Step} time includes the initialization, solving of the optimization problem, and executing Algorithm~\ref{alg:condensing} to increment the intermediate target $\hat{y}_t^{\mathrm{d}}$.  
The computational complexity of the proposed formulation is only marginally increased compared with the MPC using the naive $L_2$ offset cost, \revised{by about $17\%$--$35\%$. For both approaches and both environments, the worst-case per-step computation times are significantly below the sampling time of \SI{50}{\milli\second}, thus indicating real-time capability.}

For both MPC schemes, \revised{the computation times increase roughly linearly with the number of obstacles $n_{\mathrm{o}}$}. 
This is due to the handling of obstacle constraints (Sec.~\ref{sec:efficient_obstacle}), which increases the number of decision variables and constraints in the optimization problem. 
\revised{As a result, a direct application of the proposed algorithm to problems with a large number of obstacles, say $n_{\mathrm{o}}\geq 100$, would yield a problematic computational demand.
In practice, this can be addressed by selecting a subset of obstacles to pass to the MPC formulation, similar to~\cite{narkhede2022sequential}. 
Specifically, at any given point in time, most obstacles are not within reach of the robot over the considered prediction horizon of $\SI{1}{\second}$. 
However, we have not implemented this in the current open-source framework, as we lack a physical setup that would allow us to effectively test environments with hundreds of obstacles.} 

The \textit{Preprocessing} is a one-time cost of constructing the shortest path roadmap, given the location of the obstacles. The \textit{New Target} computes the shortest path whenever the target $\B{Y}^{\mathrm{d}}_t$ changes. The \textit{Preprocessing} and \textit{New Target} computations needed for the proposed method have a negligible computational load. 
These results suggest that the controller can deal with rapidly changing targets, which we also consider in the hardware experiments. 
From the computational demand and numerical implementation, the approach could also deal with online changing obstacle locations, however, the derived theoretical guarantees (Thm.~\ref{thm:main}/\ref{thm:MPC_eff}) are only valid for static known obstacles. 
\begin{table}[h]      
\begin{center}
\caption{\textrm{\revised{Worst-case Computation times measured across all time steps and all trials }in milliseconds of the proposed MPC with the segment-based offset cost (Proposed) and a standard MPC using the $L_2$-norm as an offset cost ($L_2$) for the sparse ($n_{\mathrm{o}}=6$) and dense ($n_{\mathrm{o}}=15$) environments.}}
\revised{\begin{tabular}{l|c|c c}
    \toprule
    & Per Step & Preprocessing & New Target \\
    \midrule
    \multicolumn{4}{l}{Sparse environment \textbf{($n_{\mathrm{o}} = 6$)}} \\
    $L_2$                 & $10.63$     & -               & -              \\
    Proposed              & $14.37$     & $0.32$ & $0.08$ \\
    \midrule
    \multicolumn{4}{l}{Dense environment (\textbf{$n_{\mathrm{o}} = 15$)}} \\
    $L_2$                 & $35.67$    & -               & -               \\
    Proposed              & $41.83$    & $2.52$ & $0.30$ \\
    \bottomrule
\end{tabular}}
    \label{tab:sim_computation}
\end{center}
\end{table}

\subsection{Hardware experiments}  
\label{sec:experiments_hardware}
 \begin{figure}[t]
    \centering
    \includegraphics[width=0.485\textwidth]{Fig/arc_screenshot_anoted_V3.png}
        \includegraphics[width=0.485\textwidth]{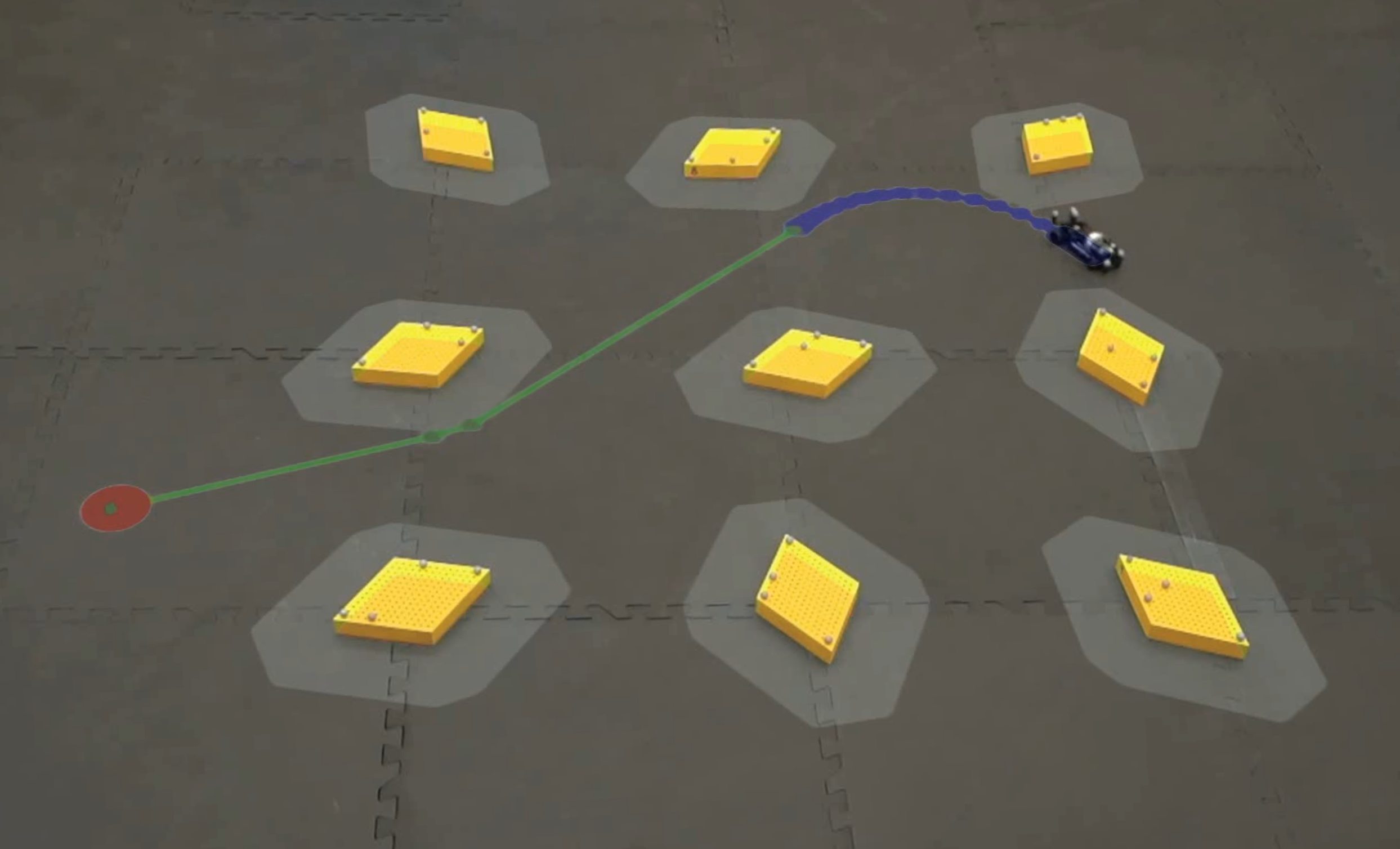}
    \caption{Snapshots from hardware experiments using two different obstacle setups.
    The target (red circle) is controlled by an operator using a drag-and-drop interface. The obstacles are yellow rhombuses, and their buffer zone is overlaid on top. 
    The MPC jointly optimizes a dynamic trajectory (blue) and a $\nu=3$-segment path (green) to reach the  target (red) or the intermediate target (purple).} 
    \label{fig:hardware}
\end{figure}
\begin{figure}[t]
    \centering
        \includegraphics[width=0.485\textwidth]{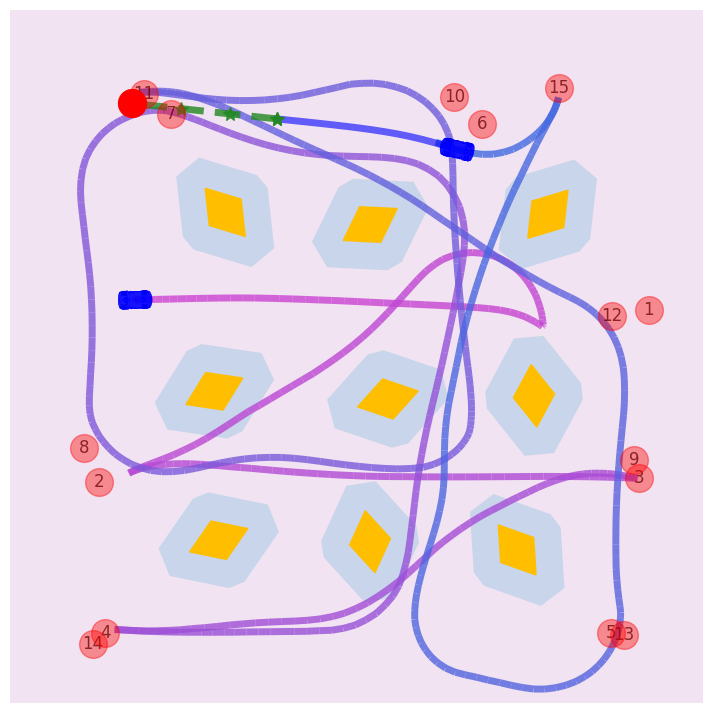}
    \caption{Closed-loop trajectory of hardware experiment, tracking $16$ different targets $\B{Y}^{\mathrm{d}}_t$ (shown in red, numbered historically) within $\SI{24}{\s}$. Car is shown at initial and final  position in blue.}
    \label{fig:hardware_v2}
\end{figure}

 
As a proof-of-concept, we demonstrated the proposed controller in hardware experiments. 
We used custom-built 1:28 miniature R/C cars based on a Mini-Z MB010 four-wheel drive chassis~\cite{carron2023CRS}. 
The vehicle’s state and the location of the obstacles are
measured using the motion capture system from Qualisys. 
We demonstrated the performance of the proposed MPC in two environments, as shown in Figure~\ref{fig:hardware}. 
One environment has $n_{\mathrm{o}}=10$ obstacles
arranged in the shape of an arc, which demonstrates the controller's ability to avoid a local minimum. 
In the second environment, $n_{\mathrm{o}}=9$ obstacles are uniformly spaced across the domain. 
In both experiments, a human operator dynamically changed the target $\B{Y}^{\mathrm{d}}_t$ using a drag-and-drop interface while the car is running. 
Snapshots of the experiments can be seen in Figure~\ref{fig:hardware}. 
Figure~\ref{fig:hardware_v2} shows the closed-loop trajectory of one of the hardware experiments. The robot moves between $16$ different targets $\B{Y}^{\mathrm{d}}_t$ in an experiment that only lasts $\SI{24}{\s}$, highlighting that the method is extremely fast in moving to a new target specified during operation\revised{, i.e., while the robot is still in motion}.  
Videos of the experiments are available online: 
\begin{center}
{\small \url{https://youtu.be/Hn_hpAmGgq0}}
\end{center}
In both demonstrations, the car was able to react to dynamic target changes, reaching new targets within 2-3 seconds.  
Furthermore, the videos show how the obstacle avoidance formulation allows us to tightly navigate obstacles -  reliably avoiding them while navigating closely using the exact geometrical description of the robot and obstacle. 
\revised{In all hardware experiments, the computations were performed within the sampling period and the control input was updated every $\SI{50}{\milli\second}$.} 
\revised{Given the control frequency and the fact that control and planning are done concurrently, the robot changes motor commands within  $\SIrange{50}{100}{\milli\second}$ after receiving a new target. 
 In contrast, if we would have utilized a classical  motion planning pipeline, then we would first have to solve a new optimization problem to determine a trajectory to the new desired target. As a result, the robot would typically first have to slow down and come to stand still before responding with new torque commands after a few seconds.}\color{black}

\section{Conclusion} 
\label{sec:conclusion}
We have presented an MPC formulation for efficient  navigation of mobile robots in environments cluttered with obstacles.
By unifying a finite-horizon trajectory optimization and a finite-segment shortest path planner, the proposed formulation ensures stability, convergence, collision avoidance, and can be implemented efficiently.
We demonstrate the applicability of the proposed method in simulation and hardware experiments using a small mobile robot. In particular, a human operator dynamically changed the target. \revised{The robot navigated through obstacles and reached new targets within  $2$--$3$ seconds and reacts to new targets within \SIrange{50}{100}{\milli\second}}. 

\revised{Future work focuses on extending this work to dynamic moving obstacles and online estimation of the environment.}

\section*{ACKNOWLEDGMENT}
We thank Shengjie Hu for cleaning up the code and simplifying the use of CRS implementation and Sabrina Bodmer for help in debugging the code in initial hardware experiments. 
 
\appendix
\subsection{Proof - Proposition~\ref{prop:offset_properties}}
\begin{proof}
\textbf{Part I. }Assumption~\ref{assump:link} ensures that Problem~\eqref{eq:offset_cost_segment} is feasible for any $y^s\in \B{S}_{\mathrm{y},\B{O}}$ given that there exists $y_{n_s}\in \B{Y}^{\mathrm{d}}_t\cap \B{S}_{\mathrm{y},\B{O}}$  (\cref{assump:feasible_target}). 
Taking into account the finite number of $n_s$ segments, the non-negative cost, and the compact set $\B{S}_{\mathrm{y}}$, the function $T_{\B{Y}_t^{\mathrm{d}}}(y^s)$ is well defined, non-negative, and uniformly bounded for all $y^s\in \B{S}_{\mathrm{y},\B{O}}$. \\
\textbf{Part II. }The cost in~\eqref{eq:offset_cost_segment} is non-increasing if we remove the constraints~\eqref{eq:offset_cost_segment_space_obstacle} and the optimal solution to the unconstrained problem is the (scaled) point-to-set distance $k_{\mathrm{M}}\|y^{\mathrm{s}}\|_{\B{Y}_t^{\mathrm{d}}}$, i.e., the second inequality in~\eqref{eq:offset_upper_bound} holds. 
Furthermore, Assumption~\ref{assump:unique} ensures
\begin{align}
\|r^{\mathrm{s}}\|_{\B{Z}_t^{\mathrm{d}}}=\min_{r^{\mathrm{d}}\in\B{Z}_t^{\mathrm{d}}}\|r^{\mathrm{s}}-r^{\mathrm{d}}\|\stackrel{\eqref{eq:assump:unique}}{\leq} k_{\mathrm{Y}}\min_{y^{\mathrm{d}}\in\B{Y}_t^{\mathrm{d}}}\|y^{\mathrm{d}}-y^{\mathrm{s}}\|,
\end{align}
which implies the first inequality in~\eqref{eq:offset_upper_bound}.\\
\textbf{Part III. }
Problem~\eqref{eq:offset_cost_segment} determines a $n_s$-segment path from $y^{\mathrm{s}}$ to $\B{Y}_t^{\mathrm{d}}$ of length $T_{\B{Y}_t^{\mathrm{d}}}(y^{\mathrm{s}})$. By following this path for a distance of $\epsilon\cdot  T_{\B{Y}_t^{\mathrm{d}}}(y^{\mathrm{s}})$, $\epsilon\in[0,1]$, we end up at a position $\hat{y}^{\mathrm{s}}\in\B{S}_{\mathrm{y},\B{O}}$ that satisfies~\eqref{eq:offset_epsilon_1} with equality. 
Furthermore, Inequality~\eqref{eq:offset_epsilon_2} follows with
\begin{align*}
\epsilon T_{\B{Y}_t^{\mathrm{d}}}(y^{\mathrm{s}})=T_{\{\hat{y}^{\mathrm{s}}\}}\|y^{\mathrm{s}}\|\leq \|y^{\mathrm{s}}-\hat{y}^{\mathrm{s}}\|\stackrel{\eqref{eq:assump:unique}}{\geq} \dfrac{1}{k_{\mathrm{Y}}}\|r^{\mathrm{s}}-\hat{r}^{\mathrm{s}}\|,    
\end{align*}
where the last inequality used $\hat{r}^{\mathrm{s}}$ from Assumption~\ref{assump:unique}.
\end{proof}

\subsection{Proof - Proposition~\ref{prop:art_reference_decrease}}
\begin{proof}
For contradiction, suppose that     
\begin{align}
\label{eq:artificial_distance_proof_1}
 \|x(t)-x_t^{\mathrm{s},\star}\|< a\|r^{\mathrm{s},\star}_t\|_{\B{Z}_t^{\mathrm{d}}},
\end{align}
with a later specified constant $a>0$. In the following, we show that a different reference $\hat{r}^{\mathrm{s}}$ would yield a smaller cost in Problem~\eqref{eq:mpc_opt}, which yields the contradiction. 
Consider $\hat{y}^{\mathrm{s}}$, $\hat{r}^{\mathrm{s}}=(\hat{x}^{\mathrm{s}},\hat{u}^{\mathrm{s}})$ from Inequalities~\eqref{eq:offset_epsilon} in Proposition~\ref{prop:offset_properties}, i.e.,
\begin{align}
\label{eq:artificial_distance_proof_2}
T_{\B{Y}_t^{\mathrm{d}}}(\hat{y}^{\mathrm{s}})\leq (1-\epsilon)T_{\B{Y}_t^{\mathrm{d}}}(y^{\mathrm{s},\star}_t),\nonumber\\
\|r^{\mathrm{s},\star}_t-\hat{r}^{\mathrm{s}}\|\leq  \epsilon k_{\mathrm{Y}} T_{\B{Y}_t^{\mathrm{d}}}(y^{\mathrm{s},\star}_t),
\end{align}
holds with some later specified $\epsilon\in[0,1]$.
We ensure feasibility of Problem~\eqref{eq:mpc_opt} and derive an upper bound on the optimal cost with the reference $r^{\mathrm{s}}_t=\hat{r}^{\mathrm{s}}$ by ensuring $\|x(t)-\hat{x}^{\mathrm{s}}\|$ is sufficiently small and invoking the local controllability in Assumption~\ref{assump:local_controllable}. 
Let us denote by $\bar{T}\in[0,\infty)$ a uniform upper bound on the offset cost $T_{\B{Y}_t^{\mathrm{d}}}(y^{\mathrm{s}})$ for any $y^{\mathrm{s}}\in\B{S}_{\mathrm{y},\B{O}}$ and any $\B{Y}_t^{\mathrm{d}}\subseteq\B{S}_{\mathrm{y},\B{O}}$, which exists due to Proposition~\ref{prop:offset_properties}. 
The distance to the artificial reference $\hat{x}^{\mathrm{s}}$ satisfies
\begin{align}
\label{eq:artificial_distance_proof_3}
 &\|x(t)-\hat{x}^{\mathrm{s}}\|\leq \|x(t)-x^{\mathrm{s},\star}_t\|+\|x^{\mathrm{s},\star}_t-\hat{x}^{\mathrm{s}}\|\nonumber\\
\stackrel{\eqref{eq:artificial_distance_proof_1},\eqref{eq:artificial_distance_proof_2}}{<}  &a\|r_t^{\mathrm{s},\star}\|_{\B{Z}_t^{\mathrm{d}}}+\epsilon k_{\mathrm{Y}}  T_{\B{Y}_t^{\mathrm{d}}}(y^{\mathrm{s},\star}_t)\nonumber\\
\stackrel{\eqref{eq:offset_upper_bound}}{\leq} &\left(a\dfrac{ k_{\mathrm{Y}}}{k_{\mathrm{M}}}+ \epsilon k_{\mathrm{Y}}  \right)T_{\B{Y}_t^{\mathrm{d}}}(y^{\mathrm{s},\star}_t) \nonumber\\
\leq& \left(a\dfrac{k_{\mathrm{Y}}}{k_{\mathrm{M}}}+ \epsilon k_{\mathrm{Y}}  \right) \bar{T}\leq \delta,
\end{align}
where the last inequality holds by choosing $a,\epsilon>0$ sufficiently small. 
Let us denote by $V^\star_{\B{Y}_t^{\mathrm{d}}}(x(t))$ the optimal cost of Problem~\eqref{eq:mpc_opt}, given the measured state $x(t)$ and the target set $\B{Y}_t^{\mathrm{d}}$. 
Given that the stage cost $\ell$ is non-negative (Asm.~\ref{assump:stage_cost_bounds}), 
the optimal cost to Problem~\eqref{eq:mpc_opt} satisfies
\begin{align}
\label{eq:artificial_distance_proof_4}
V^\star_{\B{Y}_t^{\mathrm{d}}}(x(t))\geq T_{\B{Y}_t^{\mathrm{d}}}(y^{\mathrm{s},\star}_t).
\end{align}
Given Inequality~\eqref{eq:artificial_distance_proof_3}, we invoke Assumption~\ref{assump:local_controllable} to provide a feasible candidate solution $x_{\cdot|t},u_{\cdot|t}$, $r^{\mathrm{s}}_t=\hat{r}^{\mathrm{s}}$
to Problem~\eqref{eq:mpc_opt}, which yields the following upper bound to the optimal cost:
\begin{align}
\label{eq:artificial_distance_proof_5}
& V^\star_{\B{Y}_t^{\mathrm{d}}}(x(t))\stackrel{\eqref{eq:local_upper_bound}}{\leq } k_{\mathrm{V}}\|x(t)-\hat{x}^{\mathrm{s}}\|^2 + T_{\B{Y}_t^{\mathrm{d}}}(\hat{y}^{\mathrm{s}})\\
\stackrel{\eqref{eq:artificial_distance_proof_2},\eqref{eq:artificial_distance_proof_3}}{<} &k_{\mathrm{V}}\left(a \dfrac{k_{\mathrm{Y}}}{k_{\mathrm{M}}}+ \epsilon k_{\mathrm{Y}}  \right)^2  T_{\B{Y}_t^{\mathrm{d}}}(y^{\mathrm{s},\star}_t)^2+(1-\epsilon)T_{\B{Y}_t^{\mathrm{d}}}(y^{\mathrm{s},\star}_t)\nonumber \\
\leq &T_{\B{Y}_t^{\mathrm{d}}}(y^{\mathrm{s},\star}_t)-\left[\epsilon-k_{\mathrm{V}}\bar{T} \left(a\dfrac{k_{\mathrm{Y}}}{ k_{\mathrm{M}}}+ \epsilon k_{\mathrm{Y}}  \right)^2  \right] T_{\B{Y}_t^{\mathrm{d}}}(y^{\mathrm{s},\star}_t)\nonumber\\
\leq &  T_{\B{Y}_t^{\mathrm{d}}}(y^{\mathrm{s},\star}_t), \nonumber
\end{align}
where the last inequality holds by choosing $\epsilon>0$ and $a>0$ sufficiently small. 
Conditions~\eqref{eq:artificial_distance_proof_4} and \eqref{eq:artificial_distance_proof_5} yield a contradiction, thus showing~\eqref{eq:artificial_reference_distance}. 
\end{proof}

\subsection{Proof - Theorem~\ref{thm:main}}
\begin{proof}
\textbf{Part I. }First, note that the function $T_{\B{Y}_t^{\mathrm{d}}}(y^{\mathrm{s}})$ is well-defined for any feasible \revised{position}  $y^{\mathrm{s}}\in\B{S}_{\mathrm{y},\B{O}}$, given $\B{Y}^{\mathrm{d}}_t\cap \B{S}_{\mathrm{y},\B{O}}\neq \emptyset$ (Assumption~\ref{assump:feasible_target}) and Proposition~\ref{prop:offset_properties}. 
Hence, feasibility of Problem~\eqref{eq:mpc_opt} is independent of the possible changes in the target $\B{Y}_t^{\mathrm{d}}$, as standard in MPC formulations using artificial references~\cite{krupa2024model}.  
We show recursive feasibility of Problem~\eqref{eq:mpc_opt} using induction. 
Suppose Problem~\eqref{eq:mpc_opt} is feasible at time $t\in\B{N}$ with optimal solution $x^\star_{\cdot|t},u_{\cdot|t}^\star,r^{\mathrm{s},\star}_t$. 
A feasible candidate solution is given by shifting this sequence and appending the steady-state input, i.e.,
\begin{align}
\label{eq:MPC_proof_candidate}
x_{k|t+1}=&x_{k+1|t}^\star,~k\in\B{N}_{0:N-1},~
u_{k|t+1}=u_{k+1|t}^\star,~k\in\B{N}_{0:N-2}\nonumber \\
r^{\mathrm{s}}_{t+1}=&r^{\mathrm{s},\star}_t,
~u_{N-1|k+1}=u_t^{\mathrm{s},\star}, x_{N|k+1}=x_{N-1|k+1}=x_t^{\mathrm{s},\star},
\end{align}
is a feasible solution to Problem~\eqref{eq:mpc_opt} at time $t+1$. 
This follows from the terminal equality constraint~\eqref{eq:mpc_opt_terminal} and the fact that $r^{\mathrm{s},\star}_t\in\B{S}_{\B{O}}$ is a feasible steady-state using~\eqref{eq:steady_state_obstacle}. 
Feasibility of Problem~\eqref{eq:mpc_opt} and the constraint~\eqref{eq:mpc_opt_obstacle} with $k=0$, $u_{0|t}^\star=u(t)$, $x^\star_{0|t}=x(t)$ ensure that the closed-loop state and input satisfy the input, state, and collision avoidance constraints $\B{Z}_{\B{O}}$. \\
\textbf{Part II. }Since $\B{Y}_t^{\mathrm{d}}$ is constant, we denote the constant target sets by $\B{Y}^{\mathrm{d}}$, $\B{X}^{\mathrm{d}}$ and the optimal cost of Problem~\eqref{eq:mpc_opt} by $V^\star(x(t))$. We establish stability and convergence by showing that there exist functions $\alpha_1,\alpha_2\in\mathcal{K}_\infty$:
\begin{subequations}
\label{eq:Lyap_H_MPC}
\begin{align}
\label{eq:Lyap_H_MPC_1}
\alpha_1(\|x(t)\|_{\B{X}^{\mathrm{d}}})\leq     
V^\star(x(t))\leq& \alpha_2(\|x(t)\|_{\B{X}^{\mathrm{d}}}),\\
\label{eq:Lyap_H_MPC_2}
V^\star(x(t+1))-V^\star(x(t))\leq& -\alpha_1(\|x(t)\|_{\B{X}^{\mathrm{d}}}).
\end{align}
\end{subequations}
\textbf{Lower bound: }Given that the stage cost $\ell$ and the offset cost $T$ in~\eqref{eq:mpc_opt} are non-negative, we have
\begin{align}
\label{eq:mpc_proof_lower_1}
&V^\star(x(t))\geq \ell(x(t),u_{0|t}^\star,r^{\mathrm{s},\star}_t)
\stackrel{\eqref{eq:assump:stage_cost_bounds}}{\geq }\underline{\alpha}_\ell(\|x(t)-x^{\mathrm{s},\star}_t\|)\nonumber\\
\stackrel{\eqref{eq:artificial_reference_distance}}{\geq}&  \underline{\alpha}_\ell\left(\frac{1}{2}\left(\|x(t)-x_t^{\mathrm{s},\star}\|+a\|x_t^{\mathrm{s},\star}\|_{\B{X}^{\mathrm{d}}}\right)\right)\nonumber\\
\geq &\underline{\alpha}_\ell\left(\min\{1,a\}\frac{1}{2}\|x(t)\|_{\B{X}^{\mathrm{d}}}\right)=:\alpha_1(\|x(t)\|_{\B{X}^{\mathrm{d}}}).
\end{align}
\textbf{Upper bound: }First, we establish a local upper bound, for $\|x(t)\|_{\B{X}^{\mathrm{d}}}\leq \delta$, with $\delta>0$ according to \cref{assump:local_controllable}. 
We define a pair $(x^{\mathrm{d}},u^{\mathrm{d}})\in\B{Z}^{\mathrm{d}}$, $y^{\mathrm{d}}=h(x^{\mathrm{d}})\in\B{Y}^{\mathrm{d}}$: $\|x(t)-x^{\mathrm{d}}\|=\|x(t)\|_{\B{X}^{\mathrm{d}}}\leq \delta$. 
A feasible candidate solution is given by the artificial reference $r_t^{\mathrm{s}}=(x^{\mathrm{d}},u^{\mathrm{d}})$ and 
the local controllability condition (\cref{assump:local_controllable}), ensuring
\begin{align}
\label{eq:mpc_proof_upper_1}
V^\star(x(t))\leq & \sum_{k=0}^{N-1}\ell(x_{k|t},u_{k|t},r^{\mathrm{d}})+T_{\B{Y}^{\mathrm{d}}}(y^{\mathrm{d}}) \nonumber \\\stackrel{\eqref{eq:local_upper_bound}
}{\leq} &
k_{\mathrm{V}}\|x(t)-x^{\mathrm{d}}\|^2=k_{\mathrm{V}}\|x(t)\|_{\B{X}^{\mathrm{d}}}^2.
\end{align}
Given $\B{Z}$ compact (\cref{assump:regular}), we have a uniformly bounded stage cost (\cref{assump:stage_cost_bounds}) and offset cost $T_{\B{Y}^{\mathrm{d}}}$ (Prop.~\ref{prop:offset_properties}). Hence, there exists a finite constant $\bar{V}>0$ that uniformly bounds $V^\star(x(t))$ for any feasible state $x(t)$.
Thus, for any feasible states satisfying $\|x(t)\|_{\B{X}^{\mathrm{d}}}\geq \delta$, the following bound holds:
\begin{align}
\label{eq:mpc_proof_upper_2}
V^\star(x(t))\leq & \bar{V}\leq \dfrac{\bar{V}}{\delta^2}\|x(t)\|_{\B{X}^{\mathrm{d}}}^2.
\end{align}
Combining the two cases in Inequalities~\eqref{eq:mpc_proof_upper_1} and \eqref{eq:mpc_proof_upper_2}, the upper bound~\eqref{eq:Lyap_H_MPC_1} holds with
$\alpha_2(c):=\max\{k_{\mathrm{V}}, \bar{V}/\delta^2\}c^2$, 
$\alpha_2\in\mathcal{K}_\infty$. \\
\textbf{Decrease condition: }We use the feasible candidate solution from Part I of the proof to derive an upper bound for the optimal cost at time $t+1$:
\begin{align}
\label{eq:Lyap_decrease}
&V^\star(x(t+1))-V^\star(x(t))\\
\leq&
\sum_{k=0}^{N-1}\ell(x_{k|t+1},u_{k|t+1},r^{\mathrm{s}}_{t+1})+T_{\B{Y}^{\mathrm{d}}}(y^{\mathrm{s}}_{t+1})\nonumber\\
&-\sum_{k=0}^{N-1}\ell(x^\star_{k|t},u^\star_{k|t},r^{\mathrm{s},\star}_t)-T_{\B{Y}^{\mathrm{d}}}(y^{\mathrm{s},\star}_t)\nonumber\\
=&\ell(x_{N|t}^\star,u_{N-1|t+1},r^{\mathrm{s},\star}_t)-\ell(x_{0|t}^\star,u_{0|t}^\star,r^{\mathrm{s},\star}_t)\nonumber\\
\stackrel{\eqref{eq:assump:stage_cost_bounds}}{\leq}& -\underline{\alpha}_\ell(\|x(t)-x_t^{\mathrm{s},\star}\|)
\stackrel{\eqref{eq:mpc_proof_lower_1}}{\leq} -\alpha_1(\|x(t)\|_{\B{X}^{\mathrm{d}}}),\nonumber
\end{align}
where the penultimate inequality used $x^\star_{0|t}=x(t)$~\eqref{eq:mpc_opt_initial} and 
$\ell(x_{N|t}^\star,u_{N-1|t+1},r^{\mathrm{s},\star}_t)=\ell(x_t^{\mathrm{s},\star},u_t^{\mathrm{s},\star},r_t^{\mathrm{s},\star})=0$ using~\eqref{eq:MPC_proof_candidate} and \eqref{eq:assump:stage_cost_bounds}.\\
\textbf{Stability \& Convergence: }Inequalities~\eqref{eq:Lyap_H_MPC} ensure that $V^\star$ is a Lyapunov function for the set $\B{X}^{\mathrm{d}}$ and there exists a function $\beta\in\C{KL}$: 
$\|x(t)\|_{\B{X}^{\mathrm{d}}}\leq \beta(\|x(0)\|_{\B{X}^{\mathrm{d}}},t)$, see~\cite[Thm.~B.18]{rawlings2017model}.
Convergence of $y(t)=h(x(t))$ to $\B{Y}^{\mathrm{d}}$ follows from continuity of $h$ (\cref{assump:regular}) and the definition of $\B{X}^{\mathrm{d}}$~\eqref{eq:steady_state_optimal}.
\end{proof}

\subsection{Proof - Theorem~\ref{thm:MPC_eff}}
\begin{proof}
The proof adapts the arguments from Theorem~\ref{thm:main} to account for the intermediate target $\hat{y}^{\mathrm{d}}_t$.\\
\textbf{Part I: }First, we establish feasibility of all optimization problems and routines called in Algorithm~\ref{alg:overall}. 
To establish feasibility of Problem~\eqref{eq:mpc_intermediate}, we consider four different cases depending on $\hat{y}_t^{\mathrm{d}}$ constant or $\B{Y}^{\mathrm{d}}_t$ constant, or $t=0$.  
Compared to Problem~\eqref{eq:mpc_opt}, Problem~\eqref{eq:mpc_intermediate} has an equivalent formulation for the collision avoidance constraints (cf. Lemma~\ref{lemma:obstacle_avoidance_dietz}), optimizes over a fewer segments $n_\nu$, and has an externally supplied intermediate target $\hat{y}^{\mathrm{d}}_t$ as a terminal condition of the optimized segments~\eqref{eq:mpc_intermediate_offset_terminal}. \\
\textit{Case (i):} Constant intermediate target $\hat{y}^{\mathrm{d}}_{t+1}=\hat{y}^{\mathrm{d}}_t$. Recursive feasibility of Problem~\eqref{eq:mpc_intermediate} follows with the same candidate solution from Theorem~\ref{thm:main}. \\
\textit{Case (ii): }$\B{Y}^{\mathrm{d}}_{t+1}=\B{Y}^{\mathrm{d}}_{t}$ is constant and $\hat{y}^{\mathrm{d}}_{t+1}\neq \hat{y}^{\mathrm{d}}_t$ is updated through Algorithm~\ref{alg:condensing}. In this case, Algorithm~\ref{alg:condensing} also determines the $\nu$-segment path $\bar{y}_{\cdot|t+1}$, which is a feasible candidate sequence for Problem~\eqref{eq:mpc_intermediate} with $\bar{y}_{0|t+1}=\bar{y}^\star_{0|t}$ and $\bar{y}_{n_\nu|t+1}=\hat{y}^{\mathrm{d}}_t$. \\
\textit{Case (iii): }Changing target $\B{Y}^{\mathrm{d}}_{t+1}\neq \B{Y}^{\mathrm{d}}_t$. 
The shortest road map is used to compute a path from $\bar{y}_{t}^{\mathrm{s},\star}\in\B{S}_{\mathrm{y},\B{O}}$ to $\B{Y}^{\mathrm{d}}_{t+1}$. 
As explained in Section~\ref{sec:efficient_roadmap}, the shortest path satisfies the following constraints
\begin{subequations}
\begin{align}  
&\mathscr{Y}_0(t+1)=\bar{y}^{\mathrm{s},\star}_t,\quad 
\mathscr{Y}_{n_\mathscr{Y}(t+1)}(t+1)\in\B{Y}^{\mathrm{d}}_{t+1},\\
&\overline{\mathscr{Y}_0(t+1)\mathscr{Y}_{1}(t+1)}\in\B{S}_{\mathrm{y},\B{O}},\\
&\overline{\mathscr{Y}_j(t+1)\mathscr{Y}_{j+1}(t+1)}\in\bar{\B{S}}_{\mathrm{y},\B{O}},~j\in\B{N}_{1:N_{\mathscr{Y}}(t+1)-1},
\end{align}
\end{subequations}
where the first segment is subject to slightly more relaxed constraints. 
Feasibility of Problem~\eqref{eq:mpc_intermediate} with $\bar{y}^{\mathrm{s},\star}_t$ at time $t$ and the modified 
\cref{assump:link,assump:feasible_target} considered in Theorem~\ref{thm:MPC_eff} ensure that such a path exists with length of at most $n_{\mathscr{Y}(t+1)}\leq n_{\mathrm{s}}+n_\nu$. 
Then, the first $n_\nu$ segments are moved to initialize $\bar{y}$. 
Hence, all remaining segments in $\mathscr{Y}_j(t+1)$ lie in the restricted set $\bar{\B{S}}_{\mathrm{y},\B{O}}$. 
This yields segments $\bar{y}_{\cdot|t+1}$ and an intermediate target $\hat{y}^{\mathrm{d}}_{t+1}$ that provide a feasible candidate solution to Problem~\eqref{eq:mpc_intermediate}. \\
\textit{Case (iv): } Lastly, we need to consider $t=0$.
A feasible path $\mathscr{Y}(0)$ is constructed starting from $\bar{y}^{\star\mathrm{s}}_t=y(0)\in\bar{\B{S}}_{\mathrm{y},\B{O}}$ as in case (iii). 
Furthermore, feasibility of Problem~\eqref{eq:mpc_intermediate} at $t=0$ follows since the initial condition $x(0)$ is assumed to be a feasible steady-state, i.e., $u_{k|0}=u^{\mathrm{s},\star}_{-1}$, $x_{k|0}=x^{\mathrm{s},\star}_{-1}=x(0)$ is a feasible state and input trajectory. \\
Lastly, constraint satisfaction follows analogous to Theorem~\ref{thm:main} from feasibility and the constraints posed in Problem~\eqref{eq:mpc_intermediate}. \\
\textbf{Part II: }In the following, we show convergence for a constant target $\B{Y}^{\mathrm{d}}$. 
The main challenge is that we use the intermediate target $\hat{y}^{\mathrm{d}}_t$, which is not optimized in the MPC, and hence Proposition~\ref{prop:art_reference_decrease} does not apply. 
Let us denote the optimal cost of Problem~\eqref{eq:mpc_intermediate} by $V^\star(x,\hat{y}^{\mathrm{d}})$. 
For the following analysis, we study the augmented cost 
\begin{align}
\label{eq:augmented_cost_including_shortest_path}
\bar{V}^\star(x,\hat{y}^{\mathrm{d}},\mathscr{Y}):=V^\star(x,\hat{y}^{\mathrm{d}})+k_M\sum_{j=0}^{n_{\mathscr{Y}}-1}\|\mathscr{Y}_{j+1}-\mathscr{Y}_j\|+ \epsilon\cdot n_{\mathscr{Y}},
\end{align}
where we added the offset cost of the not optimized path segments $\mathscr{Y}$ and 
a small factor $\epsilon>0$ to simplify the proof. 
First, note that incrementing the intermediate target with Algorithm~\ref{alg:condensing} yields the following decrease in $\bar{V}^\star$: 
\begin{align}
\label{eq:Lyap_decrease_prelim}
&\bar{V}^\star(x(t),\hat{y}^{\mathrm{d}}_{t+1},\mathscr{Y}(t+1))\nonumber\\
\leq &  \bar{V}^\star(x(t),\hat{y}^{\mathrm{d}}_{t},\mathscr{Y}(t))-\epsilon|n_{\mathscr{Y}(t)}-n_{\mathscr{Y}(t+1)}|.
\end{align}
In particular, $\|\mathscr{Y}_{j+2}-\mathscr{Y}_j\|\leq \|\mathscr{Y}_{j+2}-\mathscr{Y}_{j+1}\|+\|\mathscr{Y}_{j+1}-\mathscr{Y}_j\|$ using the triangular inequality and thus removing a segment does not increase the length of the path. Furthermore, the segment length $\|\mathscr{Y}_1(t)-\mathscr{Y}_0(t)\|$ is simply moved from $\mathscr{Y}(t)$ to $\bar{y}_{\cdot|t+1}$, which does not change the overall length of the path in~\eqref{eq:mpc_intermediate}.
Lastly, the number of segments $n_{\mathscr{Y}(t)}$ is non-increasing. 
Similar to Part III in the proof of Theorem~\ref{thm:main}, the standard candidate solution at time $t+1$ implies
\begin{align}
\label{eq:Lyap_decrease_intermediate}
&\bar{V}^\star(x(t+1),\hat{y}^{\mathrm{d}}_{t+1},\mathscr{Y}(t+1))-\bar{V}^\star(x(t),\hat{y}^{\mathrm{d}}_t,\mathscr{Y}(t)) \nonumber\\
\stackrel{\eqref{eq:Lyap_decrease},\eqref{eq:Lyap_decrease_prelim}}{\leq} & -\alpha_\ell(\|x(t)-x^{\mathrm{s},\star}_t\|)-\epsilon|n_{\mathscr{Y}(t)}-n_{\mathscr{Y}(t+1)}|.
\end{align} 
\textbf{Part III:} 
In the following, we show that there exists some $\alpha_4\in\mathcal{K}_\infty$, such that 
\begin{align}
\label{eq:Lyap_decrease_intermediate_v2}
&\bar{V}^\star(x(t+1),\hat{y}^{\mathrm{d}}_{t+1},\mathscr{Y}(t+1))-\bar{V}^\star(x(t),\hat{y}^{\mathrm{d}}_t,\mathscr{Y}(t)) \nonumber\\
\leq& -\alpha_4(\|x(t)\|_{\B{X}^{\mathrm{d}}}).
\end{align}
The key difference to Theorem~\ref{thm:main} is that we cannot directly apply Inequality~\eqref{eq:artificial_reference_distance} from Proposition~\ref{prop:art_reference_decrease}. 
Instead, our proof uses a case distinction based on the optimized offset cost:
\begin{align}
T(\hat{y}^{\mathrm{d}}_t,y^{\mathrm{s},\star}_t):=k_{\mathrm{M}}  \sum_{j=0}^{n_\nu-1}\|\bar{y}_{j+1|t}^\star-\bar{y}_{j|t}^\star\|.
\end{align}
\textit{Case (i): }Suppose $\hat{y}^{\mathrm{d}}_t\in \B{Y}^{\mathrm{d}}$, in which case $n_{\mathscr{Y}(t)}=0$. 
In this case, the arguments from Proposition~\ref{prop:art_reference_decrease} directly apply, since the optimized segments end in the target set, we just consider a smaller number of segments $n_\nu\leq n_{\mathrm{s}}$. Thus,  Inequality~\eqref{eq:artificial_reference_distance} holds with some $a>0$. \\
\textit{Case (ii): }Suppose $n_{\mathscr{Y}(t)}>0$ and 
$T(\hat{y}^{\mathrm{d}}_t,y^{\mathrm{s},\star}_t)\geq \delta_\epsilon$. In this case, the arguments from Proposition~\ref{prop:art_reference_decrease} also apply, yielding a potentially smaller constant $a>0$ satisfying~\eqref{eq:artificial_reference_distance}. 
In particular, the proof of Proposition~\ref{prop:art_reference_decrease} relies on the fact that we can decrease the overall offset cost by some factor that is larger than $k_V\|x(t)-\hat{x}^{\mathrm{s}}\|^2$, see~\eqref{eq:artificial_distance_proof_5}. 
For any $\delta_\epsilon>0$, we can find a sufficiently small $a,\epsilon>0$, such that these arguments remain valid and thus Inequality~\eqref{eq:artificial_reference_distance} also holds with some $a>0$.\\
\textit{Case (iii): }
Suppose $T(\hat{y}^{\mathrm{d}}_t,y^{\mathrm{s},\star}_t)\leq \delta_\epsilon$ and $n_{\mathscr{Y}(t)}>0$. 
This implies $\|\bar{y}_{2|t}^\star-\bar{y}_{0|t}^\star\|\leq \delta_\epsilon$. 
Thus, the construction of the inflated obstacle~\eqref{eq:obstacle_inflated} ensures that Algorithm~\ref{alg:condensing} updates the intermediate target $\hat{y}^{\mathrm{d}}_t$ using the optimized path $\mathscr{Y}(t+1)$. Hence, the cost $\bar{V}^\star$ decreases at least by $\epsilon|n_{\mathscr{Y}(t)}-n_{\mathscr{Y}(t+1)}|\geq \epsilon$ using Inequality~\eqref{eq:Lyap_decrease_intermediate}. \\
\textit{Combined case: }
For case (i)--(ii), Inequality~\eqref{eq:artificial_reference_distance} holds with some $a>0$. Thus, Inequality \eqref{eq:Lyap_decrease_intermediate} implies~\eqref{eq:Lyap_decrease_intermediate_v2} with some $\alpha_4\in\mathcal{K}_\infty$, analogous to the proof of Theorem~\ref{thm:main}. 
In case (iii), the cost decreases by $\epsilon>0$, which also implies~\eqref{eq:Lyap_decrease_intermediate_v2} if $\max_{(x,u)\in\B{Z}}\alpha_4(\|x\|_{\B{X}^{\mathrm{d}}})\leq \epsilon$. 
A sufficiently small $\alpha_4\in\mathcal{K}_\infty$ can always be chosen given that $\B{Z}$ is compact (Assumption~\ref{assump:regular}). 
Lastly, by picking the minimum over two the two functions, we get a function $\alpha_4\in\mathcal{K}_\infty$ satisfying~\eqref{eq:Lyap_decrease_intermediate_v2} for all cases. 
Inequality~\eqref{eq:Lyap_decrease_intermediate_v2} ensures that $\lim_{t\rightarrow\infty}\|x(t)\|_{\B{X}^{\mathrm{d}}}=0$, which implies $\lim_{t\rightarrow\infty}\|y(t)\|_{\B{Y}^{\mathrm{d}}}$ using continuity (Assumption~\ref{assump:regular}). 
\end{proof}

\bibliographystyle{IEEEtran}
\bibliography{Literature}

\input{bio}

\end{document}

%% file: bio.tex
 \begin{IEEEbiography}[{\includegraphics[width=1in,height=1.25in,clip,keepaspectratio]{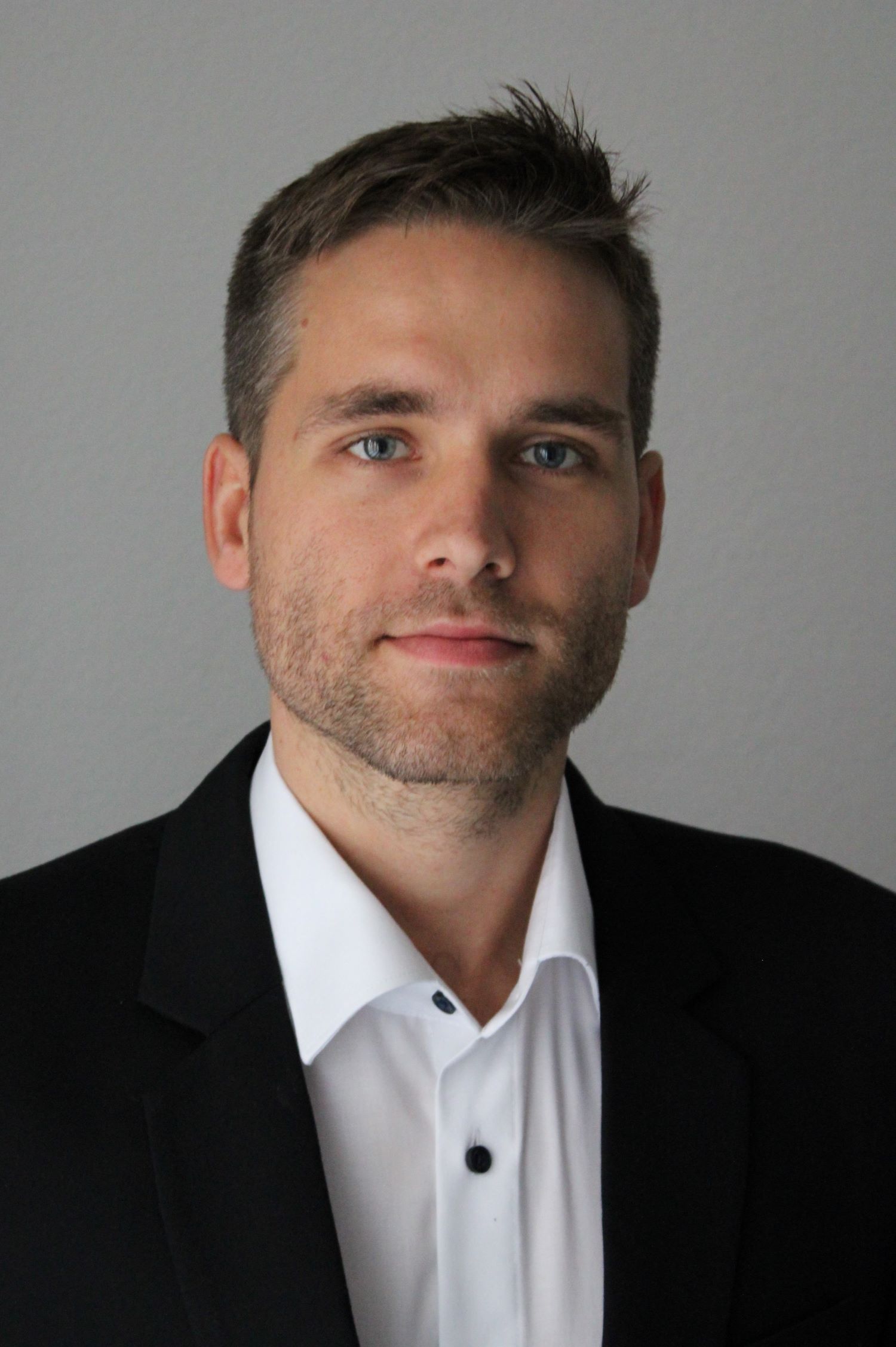}}]{Johannes K\"ohler} \looseness -1  is an Assistant Professor at Imperial College London. 
He received the Ph.D. degree from the University of Stuttgart, Germany, in 2021. 
From 2021 to 2025, he was a postdoctoral researcher at ETH Zurich, Switzerland.
He has received several awards including the 2021 European Systems \& Control PhD Thesis Award, the IEEE CSS George S. Axelby Outstanding Paper Award 2022, and the Journal of Process Control Paper Award 2023. 
His research interests include data-driven models and predictive control with applications to robotics, autonomous systems, and biomedical problems. 
\end{IEEEbiography}

\begin{IEEEbiography}[{\includegraphics[ width=.9in,height=1.25in,clip,keepaspectratio]{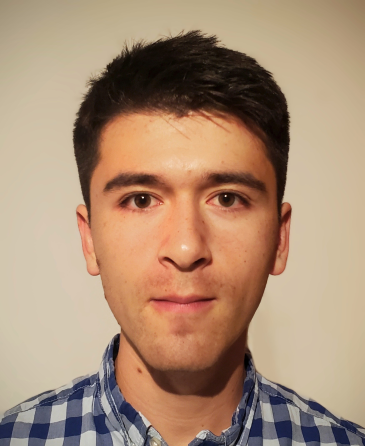}}]
{Daniel Zhang}
received his M.Sc. with distinction in Computer Science from ETH Z\"urich in 2024. During his studies, he focused on theoretical computer science, information security, and control. His research interests lie in algorithms and optimization-based control.
\end{IEEEbiography}

 \begin{IEEEbiography}[{\includegraphics[ width=1in,clip,keepaspectratio]{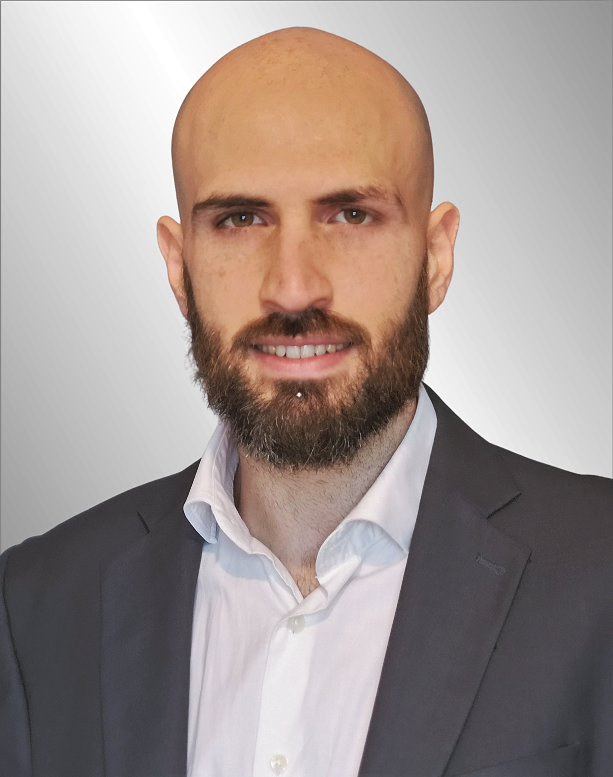}}]{Raffaele Soloperto} received his Bachelor’s and Master’s degrees in Automation Engineering from the University of Bologna, Italy, in 2014 and 2016, respectively. He earned his Ph.D. in 2022 from the University of Stuttgart, Germany, in collaboration with the International Max Planck Research School (IMPRS).
From 2022 to 2025, he was a Postdoctoral Researcher at the Automatic Control Laboratory, ETH Zürich, Switzerland.
He is currently the Innovation Manager at Embotech AG. His research interests include model predictive control and game theory.
\end{IEEEbiography}

\begin{IEEEbiography}[{\includegraphics[width=1in,height=1.25in,clip,keepaspectratio]{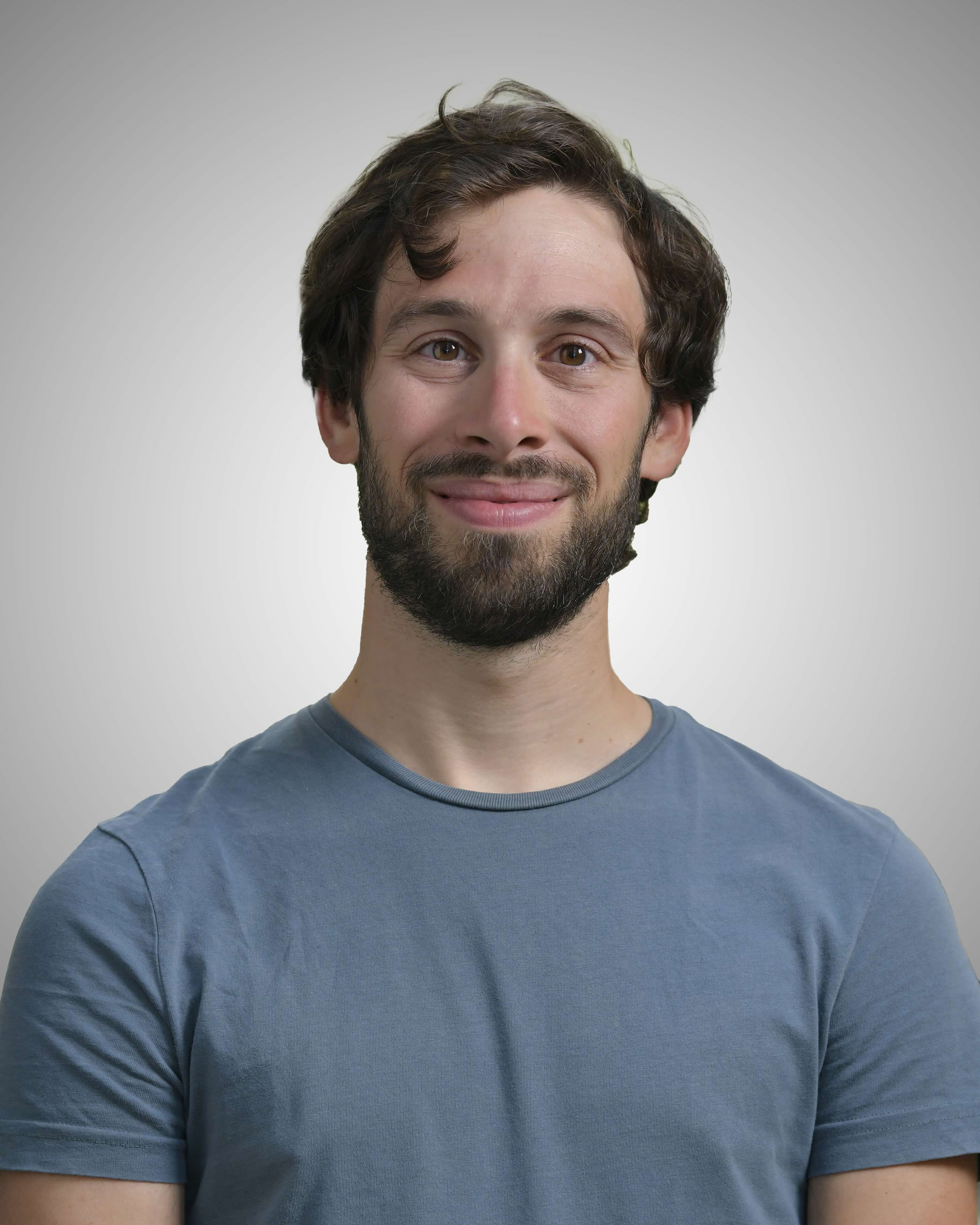}}]{Andrea Carron} received the bachelor’s, master’s, and Ph.D. degrees in control engineering from the University of Padova, Italy, in 2010, 2012, and, 2016, respectively. He is currently a Senior Lecturer with ETH Zürich. He was a Visiting Researcher with the University of California at Riverside, with Max Planck Institute in Tubingen and with the University of California at Santa Barbara, respectively. From 2016 to 2019, he was a Postdoctoral Fellow with Intelligent Control Systems Group at ETH Zürich. His research interests include safe-learning, learning-based control, multiagent systems, and robotics.
\end{IEEEbiography}

\begin{IEEEbiography}[{\includegraphics[width=1in,height=1.25in,clip,keepaspectratio]{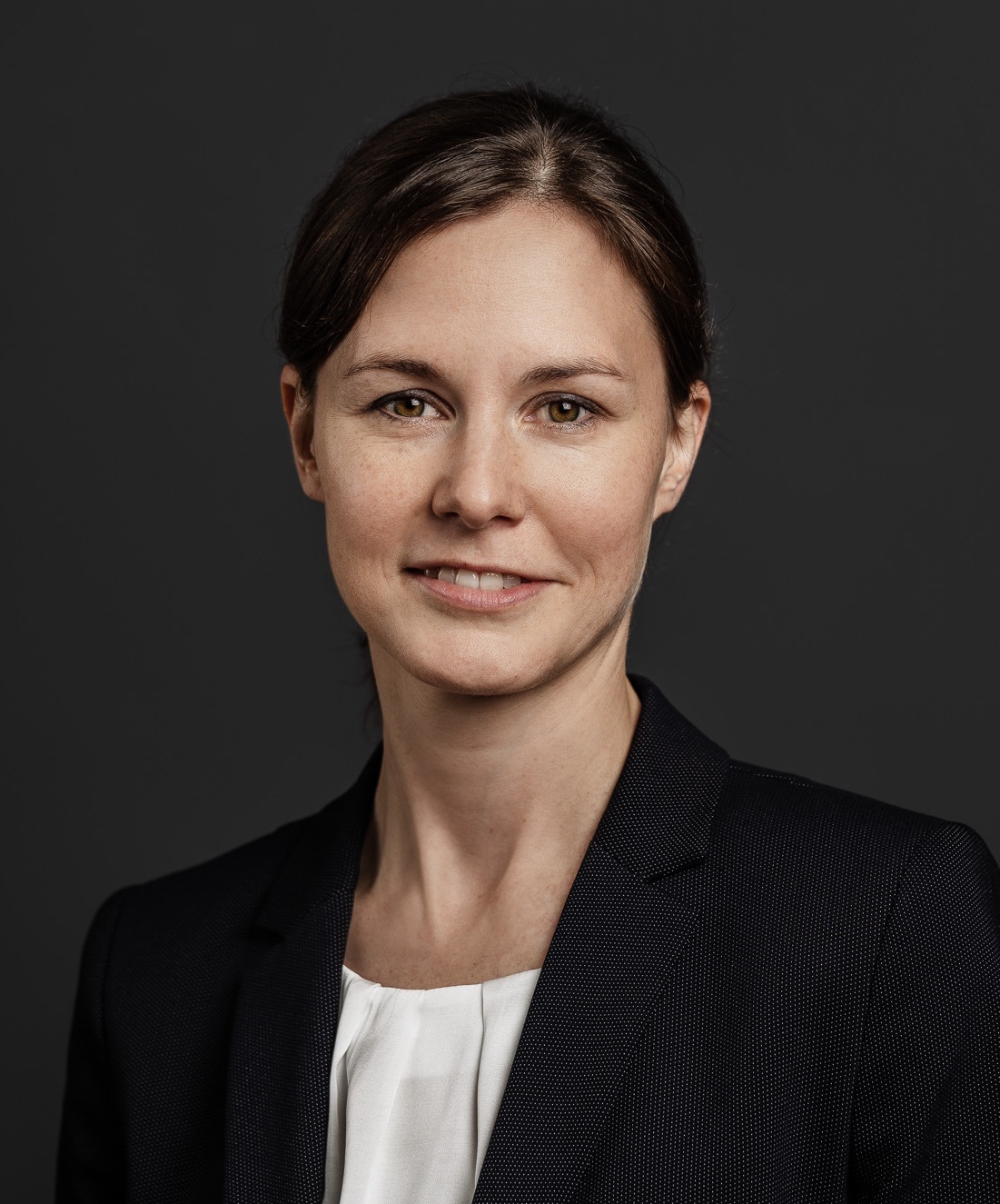}}]{Melanie N. Zeilinger}
 is an Associate Professor at ETH Zürich, Switzerland. 
She received the Diploma degree in engineering cybernetics from the University of Stuttgart, Germany, in 2006, and the Ph.D. degree with honors in electrical engineering from ETH Zürich, Switzerland, in 2011. 
From 2011 to 2012 she was a Postdoctoral Fellow with the École Polytechnique Fédérale de Lausanne (EPFL), Switzerland.
She was a Marie Curie Fellow and Postdoctoral Researcher with the Max Planck Institute for Intelligent
Systems, Tübingen, Germany until 2015 and with the Department of Electrical Engineering and Computer Sciences at the University
of California at Berkeley, CA, USA, from 2012 to 2014. 
From 2018 to 2019 she was a professor at the University of Freiburg, Germany. 
Her current research interests include safe learning-based control, as well as distributed control and optimization, with applications to robotics and human-in-the loop control.
\end{IEEEbiography}